%% file: arxiv_paper.tex
\documentclass[preprint]{article}

\usepackage{times}
\usepackage{hyperref}
\usepackage{url}

\usepackage{amsmath}
\usepackage{graphicx}
\usepackage{multicol} 
\usepackage[square,sort,comma,numbers]{natbib}
\usepackage{frcursive}
\usepackage[T1]{fontenc}

\usepackage[utf8]{inputenc} 
\usepackage[T1]{fontenc}    
\usepackage{hyperref}       
\usepackage{url}            
\usepackage{booktabs}       
\usepackage{amsfonts}       
\usepackage{nicefrac}       
\usepackage{microtype}      
\usepackage{xcolor}         

\usepackage{titling}
\usepackage{algpseudocode}

\title{Deep Generative Modeling for Identification of Noisy, Non-Stationary Dynamical Systems}

\author{
  Doris Voina \thanks{Corresponding author} \\
  \small{Department of Electrical Engineering}\\
  \small{University of Washington}\\
  \small{Seattle, WA 98195} \\
  \small{\texttt{dvoina@uw.edu}} \\
  \and
  Steven Brunton \\
  \small{Department of Mechanical Engineering} \\
  \small{University of Washington} \\
  \small{Seattle, WA 98195}\\
  \and
  J. Nathan Kutz \\
  \small{Department of Electrical Engineering}\\
  \small{University of Washington} \\
  \small{Seattle, WA 98195} \\
}
\date{\vspace{-5ex}}

\begin{document}
\maketitle

\begin{abstract}
A significant challenge in many fields of science and engineering is making sense of time-dependent measurement data by recovering governing equations in the form of differential equations. We focus on finding parsimonious ordinary differential equation (ODE) models for nonlinear, noisy, and non-autonomous dynamical systems and propose a machine learning method for data-driven system identification. While many methods tackle noisy and limited data, non-stationarity -- where differential equation parameters change over time -- has received less attention. Our method, dynamic SINDy, combines variational inference with SINDy  (sparse identification of nonlinear dynamics) to model time-varying coefficients of sparse ODEs. This framework allows for uncertainty quantification of ODE coefficients, expanding on previous methods for autonomous systems. These coefficients are then interpreted as latent variables and added to the system to obtain an autonomous dynamical model.  We validate our approach using synthetic data, including nonlinear oscillators and the Lorenz system, and apply it to neuronal activity data from C. elegans. Dynamic SINDy uncovers a global nonlinear model, showing it can handle real, noisy, and chaotic datasets. We aim to apply our method to a variety of problems, specifically dynamic systems with complex time-dependent parameters.
\end{abstract}

\section{Introduction}

Many fields of science and engineering now benefit from unprecedented amounts of data due to increased efforts and technological breakthroughs in data collection. The challenge is to use these measurements to expand our understanding of dynamical systems in areas like climate science, neuroscience, ecology, finance, and epidemiology. Machine learning methods, such as neural networks, are widely used for data-driven modeling, offering high prediction accuracy but limited interpretability. In contrast, traditional techniques that identify ordinary and partial differential equations (ODEs and PDEs) provide interpretable and generalizable insights into the system’s underlying physics. While neural networks may lose accuracy as conditions change, in many systems the governing differential equations remain reliable. The key question is whether we can combine the strengths of deep learning with the clarity and simplicity of data-driven differential equation models.
\\
\\
A key challenge in data-driven system identification is that many systems exhibit nonlinear behavior, such as switching between dynamical regimes \citep{epidemiology, security, network_grid, biophysics}. These "hybrid systems" \citep{niall_hybrid}, where continuous dynamics shift at discrete events, are more challenging to define and simulate than classical systems with smooth vector fields \citep{hybrid_hard1, hybrid_hard2}. Standard methods often assume that the data comes from a system governed by a fixed set of equations and terms, but time-varying hidden variables can further hinder identification of the system's underlying dynamics. This motivates our focus on non-autonomous (or non-stationary) systems, where sudden shifts or hidden continuous dynamics complicate accurate modeling and prediction.
\\
\\
We introduce dynamic SINDy, a data-driven method for finding non-autonomous dynamic systems with switching or continuously-varying latent variables. These systems are described by:
\begin{align}
\dot{\textbf{x}} = f(\textbf{x}(t), t)
\end{align}
where $\textbf{x}$ is vector-valued. A simple such example is $\dot{\textbf{x}} = A(t) \textbf{x}$. Importantly, we focus on systems where the time-varying component and the main variables of interest $\textbf{x}$ are separable  (e.g., $\dot{\textbf{x}} = f(\textbf{x},t) = \sin(t) \textbf{x}$, but not $\dot{\textbf{x}} = f(\textbf{x},t) = \sin(t\textbf{x})$. Another assumption is that if multiple trajectories of the system are available, these all display the same underlying switching or hidden variable dynamics.
\\
\\
Dynamic SINDy combines the interpretability of differential equations with the power of deep learning. It uses a deep generative model to uncover sparse governing equations directly from data, employing a variational autoencoder (VAE) to generate time series for differential equation coefficients. This enables data-driven discovery of equations for noisy and non-autonomous systems. The paper is organized as follows: Section 2 introduces key concepts, including SINDy, variational autoencoders, and dynamic VAEs. Section 3 describes the methodology, covering the datasets and the dynamic SINDy framework. Section 4 demonstrates dynamic SINDy's performance on various systems, including non-autonomous oscillators, Lorenz, Lotka-Volterra, and neural activity data from C. elegans. It also compares dynamic SINDy to switching linear dynamical systems \cite{slds} and group sparse regression methods \cite{group_sparsity_method}. Section 5 concludes the paper.

\section{Background and previous work}
\subsection{System identification of non-linear dynamical systems (SINDy)}
SINDy (Sparse Identification of Nonlinear Dynamics) \cite{brunton_sindy} is a data-driven method that uses sparse regression on a library of nonlinear candidate functions to match data snapshots with their derivatives, revealing the governing equations. The method assumes that only a few key terms explain the system’s dynamics. More specifically, consider $\textbf{x}(t) \in \mathbb{R}^d$ governed by the ODE: $\dot{\textbf{x}}(t) = f (\textbf{x}(t))$. Given $m$ snapshots of the system $\textbf{X} = [\textbf{x}(t_1), \textbf{x}(t_2), ..., \textbf{x}(t_m)]^T$ and the estimated time derivatives $\dot{\textbf{X}} = [ \dot{\textbf{x}}(t_1), \dot{\textbf{x}}(t_2), ..., \dot{\textbf{x}}(t_m)]^T$ , we construct a library of candidate functions $\Theta(\textbf{X}) = [1, \textbf{X}, \textbf{X}^2, ..., \textbf{X}^p, \sin(\textbf{X}), \cos(\textbf{X}), ...]$.
We then solve a sparse regression problem, $\dot{\textbf{X}} = \Theta(\textbf{X}) \Xi$, to identify the optimal coefficients $\Xi$ and to reduce the number of terms, enforcing parsimony. A sparsity-promoting regularization function $R$ is added to the final loss to yield:
\begin{align}
\hat{\Xi}= \textrm{argmin}_{\Xi} (\dot{\textbf{X}} - \Theta(\textbf{X})\Xi)^2 + R(\Xi) 
\end{align}
Several innovations have followed the original formulation of SINDy \cite{rudy_pde_group, sindy_pde, sindy_rational, sindy_corrupt}. For instance, integral and weak formulations \cite{sindy_weak, sindy_integral} have enhanced the algorithm's robustness to noise. Of relevance to our study, SINDy's generalization to non-autonomous dynamical systems has been previously explored using group sparsity norms \cite{rudy_pde_group} or clustering algorithms \cite{niall_hybrid}. 
\subsection{(Dynamic) Variational autoencoders for system identification}
The Variational Autoencoder (VAE)  \cite{vae_kingma, vae_rezende} combines neural network-based autoencoders with variational inference for probabilistic modeling and data generation. Unlike standard autoencoders, VAEs stand out due to two key features: (i) VAEs encode input data $X$ as a distribution in the latent space, allowing the decoder to generate new data by sampling from this distribution; and (ii) a regularization term ensures the latent space resembles a standard (e.g., normal) distribution, making it continuous (nearby points generate similar outputs) and complete (all points produce meaningful data). Further mathematical details can be found in Supplementary Material (SM) Section 1.1. A related method of interest is HyperSINDy \cite{hypersindy}. It combines VAEs with SINDy to discover differential equations from data. The VAE approximates the probability distribution of equation coefficients, so that once trained, HyperSINDy generates accurate stochastic dynamics and quantifies uncertainty, making it a powerful tool for model discovery.
\\
\\
In order to adapt the VAE/SINDy framework to non-autonomous systems, we would like to implement generative architectures that capture the temporal dependencies in sequential data. Dynamic VAEs (DVAEs) is an approach that extends VAEs to handle time series data \cite{timeVAE_review}. A number of DVAE architectures are described that use recurrent neural networks or state-space models to address both latent and temporal relationships \cite{dvae11_dkf, dvae12_dkf, dvae2_kva, dvae3_srn, dvae41_vrnn, dvae42_vrnn, dvae5_srnn, dvae6_rva, dvae7_dsa}. We specifically use timeVAE \cite{timeVAE}, which has shown strong performance in generating time series data by processing entire sequences with dense and convolutional layers to capture correlations. Our approach is flexible, allowing the VAE architecture to be swapped for other models better suited to the data or system under study (SM Sec. 1.2).

\subsection{Other machine learning methods for non-autonomous dynamical systems}
Traditionally, methods for handling hybrid or switching systems often involve dividing time or space into segments \cite{arma}. For instance, reduced-order models for nonlinear systems segment time intervals into smaller windows, then build a local, reduced approximation space for each segment \cite{45_time_partition1, 46_time_partition2, 47_time_partition3}. Clustering methods are also employed for modeling, particularly in complex fluid flows, where clusters represent states that can transition via a Markov model \cite{48_cluster_partition1, 49_cluster_partition} or via dynamic mode decomposition with control \cite{50_DMDc_local, 51_DMDc_local2}. Through data-informed geometry learning, authors in \cite{61_kernel_w_diffusion_map} reconstruct the relevant ``normal forms", which are prototypical realizations of the dynamics, providing bifurcation diagram and insights about the parameters even for non-autonomous systems. Yet another method \cite{43_koopman_nonautonomous} applies Koopman operator theory using DMD algorithms to find time-dependent eigenvalues, eigenfunctions, and modes in linear non-autonomous systems.
\\
\\
We compare dynamic SINDy with two existing methods (Section 4.6). First, we look at a method (recurrent SLDS) \cite{slds} that extends switching linear dynamical systems (SLDS) \cite{classical_SLDS1, classical_SLDS2} by generating transitions through changes in a continuous latent state and external inputs, rather than relying on a discrete Markov model for switching states. This model breaks the data into simpler segments and is interpretable, generative, and efficiently fitted using modular Bayesian inference. Second, we examine a method from \cite{classical_SLDS1, classical_SLDS2} that uses group-sparse penalization for model selection and parameter estimation. This method assumes shared sparsity across parameters by applying group-sparsity regularization to smaller time windows in the data, identifying the system for each segment, and then combining the results.

\section{Methods}
\subsection{Datasets}

\begin{figure}[h!]
\centering
\includegraphics[width=\textwidth]{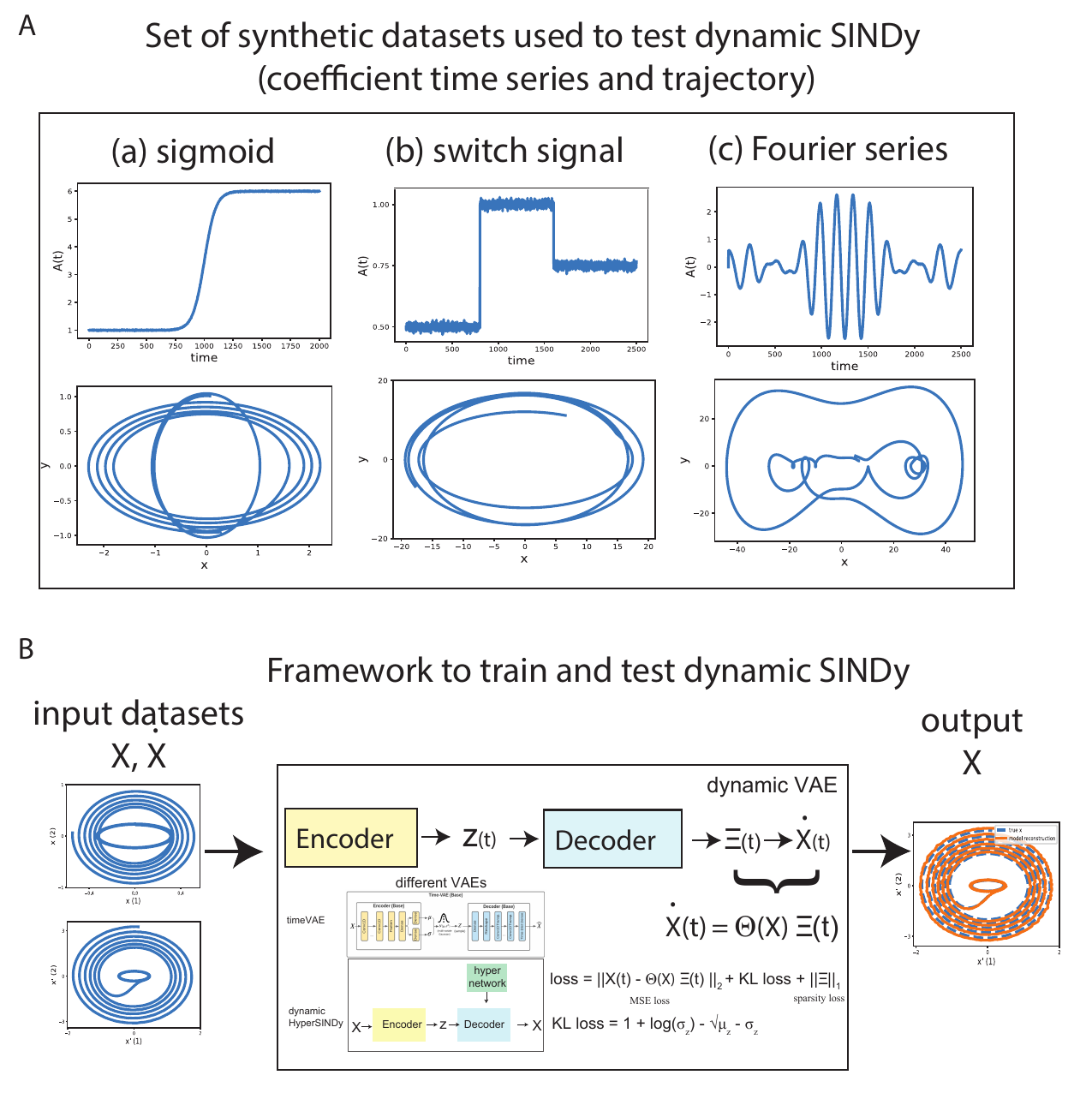}
\caption{(A). Synthetic dataset to test dynamic SINDy with non-autonomous harmonic oscillators (Eq. (\ref{eq_nonlinear_harmonic_osc})). Top: Example (SINDy) coefficient time series $A(t)$; Bottom: corresponding trajectories in phase space (B). Dynamic SINDy general architecture schematic; two DVAEs shown as example.}
\label{fig1}
\end{figure}

We use a synthetic dataset capturing dynamics of a non-autonomous harmonic oscillator:
\begin{align}
\dot{x} &= A(t) y \nonumber \\
\dot{y} &= B(t) x
\label{eq_nonlinear_harmonic_osc}
\end{align}
where $A(t)$ and $B(t)$ are the time-varying coefficients of the ODE. The time dependence of these coefficients renders the system non-autonomous and difficult to discover using classical methods. We test our approach to see if it can handle switching coefficients, as well as explore continuously varying coefficients, such as sinusoidal functions at different frequencies or finite Fourier series (Figure \ref{fig1}A, Suppl. Fig. 2). To ensure robustness against randomness, we add Gaussian noise with varying levels of variance to the time series.
\\
\\
We replace a set of constant coefficients with a set of time series (sigmoidal, switching, sinusoidal, finite Fourier series) for more complex systems, such as the chaotic Lorenz system:
\begin{align}
    \dot{x} &=\sigma(t) (y-x) \nonumber \\
\dot{y} &= x(\rho(t) -z) - y \\
\dot{z} &= xy - \beta(t) z \nonumber 
  \end{align}
We use large-scale neural recordings from whole-brain imaging to model neuronal population dynamics. C. elegans, with its 302 precisely mapped neurons, offers an ideal balance of simple behavior and complex neuronal activity. We analyze calcium imaging data from Kato et al., which includes neural recordings from the head ganglia and manual annotations of seven behaviors: forward movement, reversal, two types of reversal-to-forward turns, and two forward-to-reversal turns \cite{kato}. Previous studies show that high-dimensional neuronal activity simplifies into low-dimensional patterns, with clear clusters in principal component space representing forward and backward movements. This provides a valuable opportunity to study the link between neural activity and behavior.


\subsection{System identification for non-autonomous dynamical systems}

We explore various VAE architectures designed for inference and generation of time series data. The input is the original time series $X$, and the output are time series of ODE coefficients: 
\begin{equation}
\Xi_{1:t} = V(\textbf{X}_{1:t})
\label{xi_is_v_of_x}
\end{equation}
where $V$ is the (VAE) architecture, and $\Xi_{1:t}$ is the output time series. `Autoencoder" is a misnomer because the input is not designed to match the output in this VAE architecture. The ODE coefficients are linearly combined with a pre-determined SINDy library of basis functions to yield $\hat{\dot{\textbf{X}}}$:
\begin{equation}
\hat{\dot{\textbf{X}}}(t) = \Theta(\textbf{X}(t), t) \cdot \Xi(t)
\label{SINDy_nonautonomous}
\end{equation}
where $\Theta(\textbf{X}(t), t)$ is a row vector comprising of a polynomial basis up to cubic monomials: $\Theta(\textbf{X}(t), t) = [1 \hspace{0.1in} X_1(t) \hspace{0.1in} ... \hspace{0.1in} X_n(t)\hspace{0.1in} X_1^2(t) \hspace{0.1in} ... \hspace{0.1in} X_n^3(t)]$, where $X_i$ are features of $\textbf{X}$. Although we choose a polynomial basis for all of our experiments, the basis can change depending on the problem at hand or any prior information \cite{brunton_sindy}.  \\
Our goal is to match $\Tilde{\dot{\textbf{X}}}$, the derivative we estimate from data using numerical methods, to the output $\hat{\dot{\textbf{X}}}$ of our model (Eq. (\ref{xi_is_v_of_x}-\ref{SINDy_nonautonomous})). The loss function takes the following form:
\begin{equation}
\textrm{loss} = \sum_t || \Tilde{\dot{\textbf{X}}}(t) - \hat{\dot{\textbf{X}}}(t)|| + \lambda_1 R_{kld} + \lambda_2 R(\Xi)
\end{equation}
where $\lambda_{1,2}$ are hyperparameters of the optimization and $R, R_{kld}$ are regularization terms. $R_{kld}$ is the Kullback-Leibler divergence (KLD) loss, part of the ELBO (evidence lower bound) loss in VAEs (see SM Section 1.1). Regularization terms impose that $\Xi(t)$ is sparse (in coefficients) to encourage parsimony and that $\Xi(t)$ has minimal total variation. More details about the loss function and training, specifically the inference and generation models, can be found in the  SM, Sec. 1.3.
\\
\\
We focus on two neural network architectures in our experiments. First, timeVAE (SM Sec. 1.2.1, Suppl. Fig. 1A) is simple for proof-of-concept testing \cite{timeVAE}; however, its major drawback is that it requires the entire time series as input, which can be impractical for long sequences, especially in high-dimensional systems due to memory constraints. To address this, we introduce a new architecture called dynamic HyperSINDy (SM Sec. 1.2.2, Suppl. Fig. 1B). Alternatively, we can use DVAE architectures, which allow sequential data input, overcoming timeVAE’s limitations \cite{timeVAE_review}.

\section{Results}

\subsection{System identification of non-autonomous harmonic oscillators}

\begin{figure}[h!]
\centering
\includegraphics[width=\textwidth]{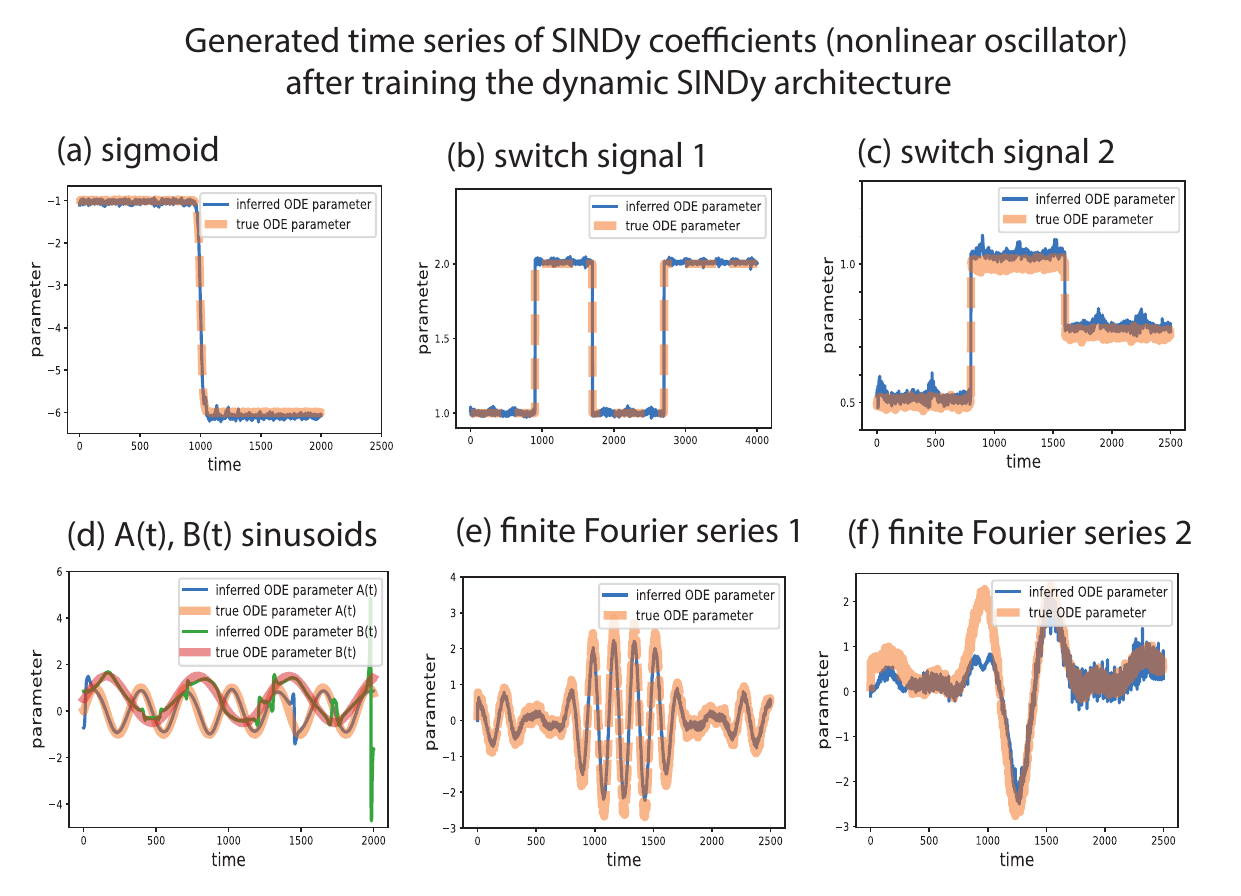}
\caption{Dynamic SINDy generates coefficient time series that match ground truth for non-autonomous harmonic oscillators (Eq. (\ref{eq_nonlinear_harmonic_osc})). (a)-(f) different examples of time-varying $A(t)$, $B(t)$.}
\label{fig2}
\end{figure}

We begin by identifying noisy, non-autonomous dynamical systems using a simple toy model -- a non-autonomous harmonic oscillator with time-varying ODE coefficients (Eq. \ref{eq_nonlinear_harmonic_osc}, Figure \ref{fig1}). First, we vary the coefficient $A(t)$ in a switch-like fashion (Figure \ref{fig2}(a)-(c)). The system behaves like a classic harmonic oscillator, but with a frequency switch. The inferred coefficients (Figure \ref{fig2}, Suppl. Fig. 4) and the reconstructed trajectories (Suppl. Fig. 6) align well with the true values. These trajectories are generated during testing, with $z$ sampled from a standard normal distribution.
\\
\\
When varying both $A(t)$ and $B(t)$ as sinusoids with different frequencies, the resulting trajectories generally capture their oscillations, though some higher error and a large outlier appear toward the end (Figure \ref{fig2}(d), Suppl. Fig. 3). We also successfully reproduce coefficients composed of multiple frequencies (a finite Fourier series) in Figures \ref{fig2}(e)-(f). In (f), some error occurs in the first half because the system approaches a fixed point where the derivative is nearly zero. In such cases, system identification becomes difficult, as multiple solutions can produce the same dynamics.

\subsection{Uncertainty quantification in non-autonomous harmonic oscillators}
\begin{figure}[h!]
\centering
\includegraphics[width=\textwidth]{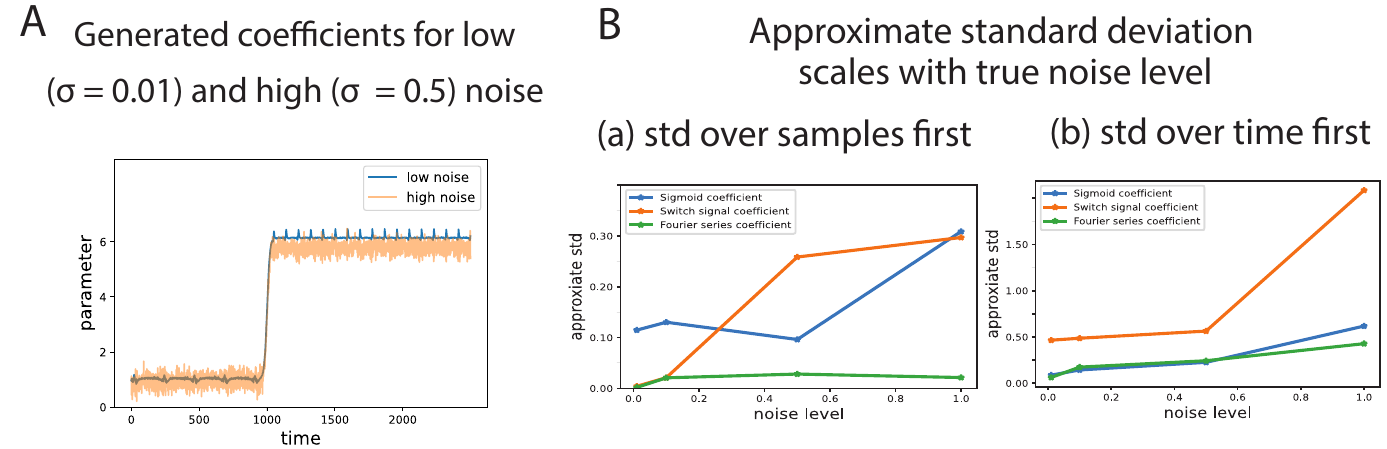}
\caption{(A) Dynamic SINDy generates coefficient time series for different levels of Gaussian noise in the coefficient. (B) Inferred noise (standard deviation, or std) scales with ground truth Gaussian noise for different time-varying coefficients. (a) std computed over many generated samples, then averaged (b) std computed over time, then averaged over samples (see Sec. 4.2)}
\label{fig3}
\end{figure}

We use VAEs to quantify uncertainty by estimating the standard deviation of the coefficients over time. Therefore we generate multiple trajectories by sampling $z$ from a standard normal distribution during testing. Figures \ref{fig3}(a)-(c) show examples of trajectories from networks trained on noisy data with two noise levels: low ($0.01$) and high ($0.5$) standard deviations. As expected, trajectories vary more under high noise than low noise. Our results show that the estimated standard deviation generally follows the true coefficient variations. First, we compute the standard deviation across generated samples at each time point and average these deviations (Figure \ref{fig3} B(a)). Second, we subtract a smooth mean from the trajectory samples and calculate the standard deviation over time (Figure \ref{fig3} B(b)). Both methods demonstrate that standard deviation aligns with the ground truth, particularly for switch signal coefficients, but is less clear for Fourier series coefficients. Further work is needed to improve standard deviation estimation, considering the VAE architecture and hyperparameters.
\subsection{System identification in a non-autonomous, chaotic toy dataset}
\begin{figure}[h!]
\centering
\includegraphics[width=\textwidth]{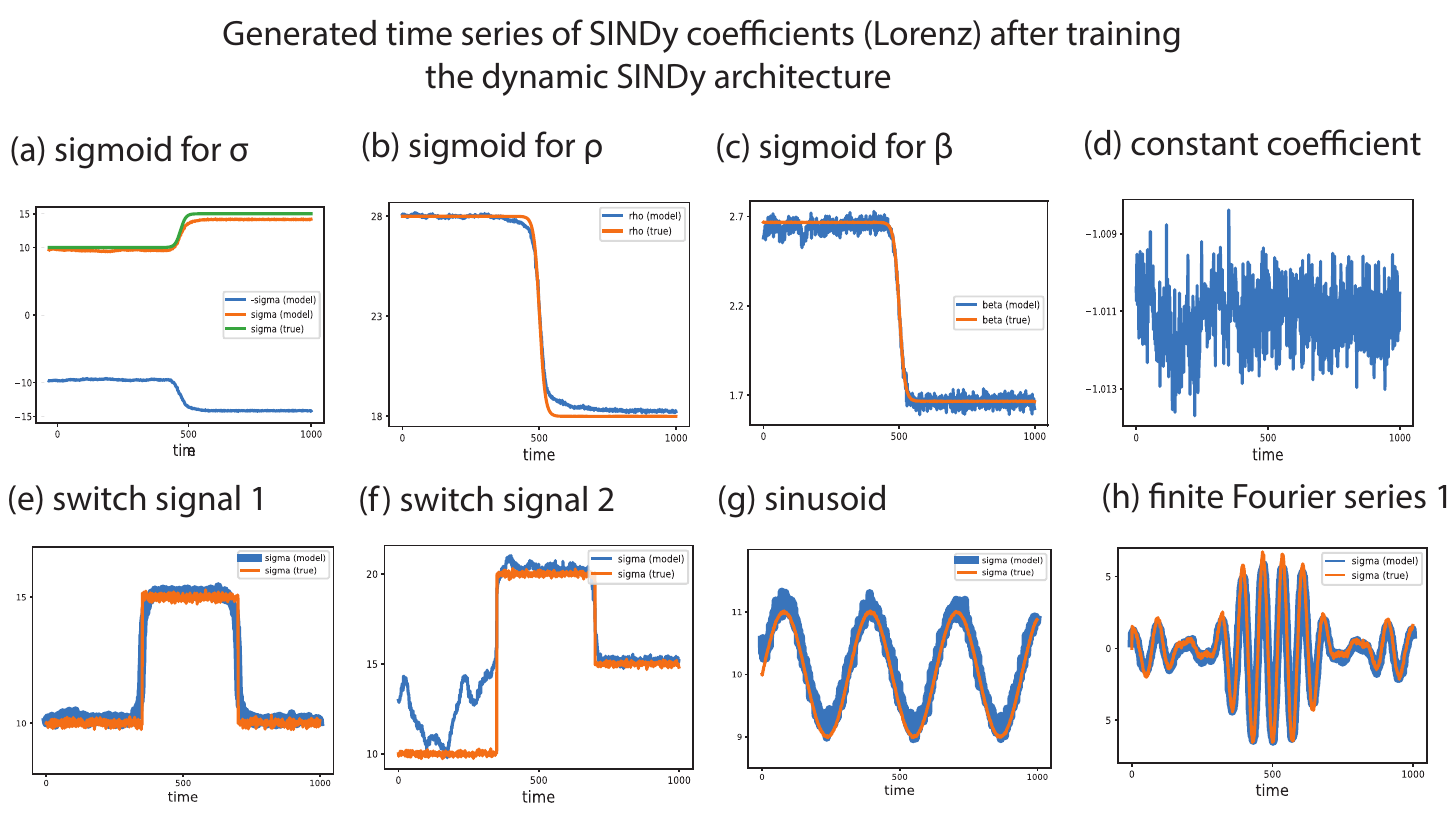}
\caption{Dynamic SINDy generates coefficient time series that match ground truth for Lorenz dynamics (Eq. (\ref{lorenz_nonautonomous})). (a)-(h) different examples of time-varying $\sigma(t)$, $\rho(t)$,  $\beta(t)$.}
\label{fig4}
\end{figure}

We next modified the Lorenz system by allowing one of its key parameters ($\sigma, \rho, \beta$) to vary over time, similar to the non-autonomous harmonic oscillator examples. The modified Lorenz equations are:
\begin{align}
\dot{x} &= \sigma(t) (y-x) \nonumber\\
\dot{y} &= x(\rho - z) - y \nonumber\\
\dot{z} &= xy - \beta z  
\label{lorenz_nonautonomous}
\end{align}
Here, $\sigma(t)$ varies over time as a sigmoid, switch function, sinusoid, or as a Fourier series with 7 overlapping frequencies. Despite these changes, the system still converges to a global attractor.
\\
\\
For system identification, we used two dynamic SINDy architectures: the timeVAE, effective for shorter time series (1000–2000 points), and dynamic HyperSINDy (SM Sec. 1.2.2), suitable for longer time series. Training occurs in two stages: first, we apply a sparsity penalty to set small coefficients to zero; second, we fine-tune the remaining coefficients. After training, we remove the encoder and generate time series from the decoder, closely matching the ground truth across different parameters and functions (Figure \ref{fig4}, Suppl. Fig. 5). Hyperparameters are listed in SM, Sec. 1.3.
\subsection{Dynamic SINDy used for identifying latent variables and their dynamics}
Dynamic SINDy is particularly useful for discovering hidden (latent) variables from incomplete datasets. We demonstrate this using a toy model from ecology: the Lotka-Volterra equations, which describe predator-prey dynamics between two species:
\begin{align}
\dot{x} &= \alpha x - \beta xy \nonumber \\
\dot{y} &= -\gamma y + \delta xy 
\label{lotka_volterra}
\end{align}
In our example, we only observe the prey population, $x$, and aim to use dynamic SINDy to uncover the hidden predator population, $y$, and reconstruct a full 2D autonomous system in $x$ and $y$.
\\
\\
We apply dynamic SINDy to $x$, using a library with just three terms: $x, x^2, x^3$. As expected, $\dot{x}$ is expressed solely in terms of $x$, with the $x^2$ and $x^3$ terms vanishing. We derive a time series for the coefficient $\tilde{y}$, where $\dot{x} = x \tilde{y}(t)$. This inferred $\tilde{y}$ correlates with the hidden $y$, where $\tilde{y} = q \cdot (\alpha - \beta y)$, with $q$ being a scaling factor applied to $x$ before using dynamic SINDy. From $\tilde{y}(t)$, we can infer $y$ and compare it to the true population. In noiseless data, we accurately reconstruct the predator dynamics \ref{fig5}A, but with more noise, recovery becomes harder \ref{fig5}B. Using $\tilde{y}$, we form a new 2D system of equations:
\begin{align}
\dot{x} &= a \cdot x \tilde{y}\nonumber \\
\dot{\tilde{y}} &= b + c \cdot x + d \cdot \tilde{y} + e \cdot x \tilde{y}
\label{lv_changed}
\end{align}
where $a,b,c,d,e$ are new model parameters. Comparing the inferred coefficients to the original Lotka-Volterra system by changing variables from $y$ to $\tilde{y}$ and using standard SINDy and the pysindy package, we find a close match (Figure \ref{fig5}C).
We applied this same approach to the non-autonomous harmonic oscillator (Eq. \ref{eq_nonlinear_harmonic_osc}, SM Sec. 2.1), further confirming that dynamic SINDy can successfully identify hidden variables and form complete autonomous systems.

\begin{figure}[h!]
\centering
\includegraphics[width=\textwidth]{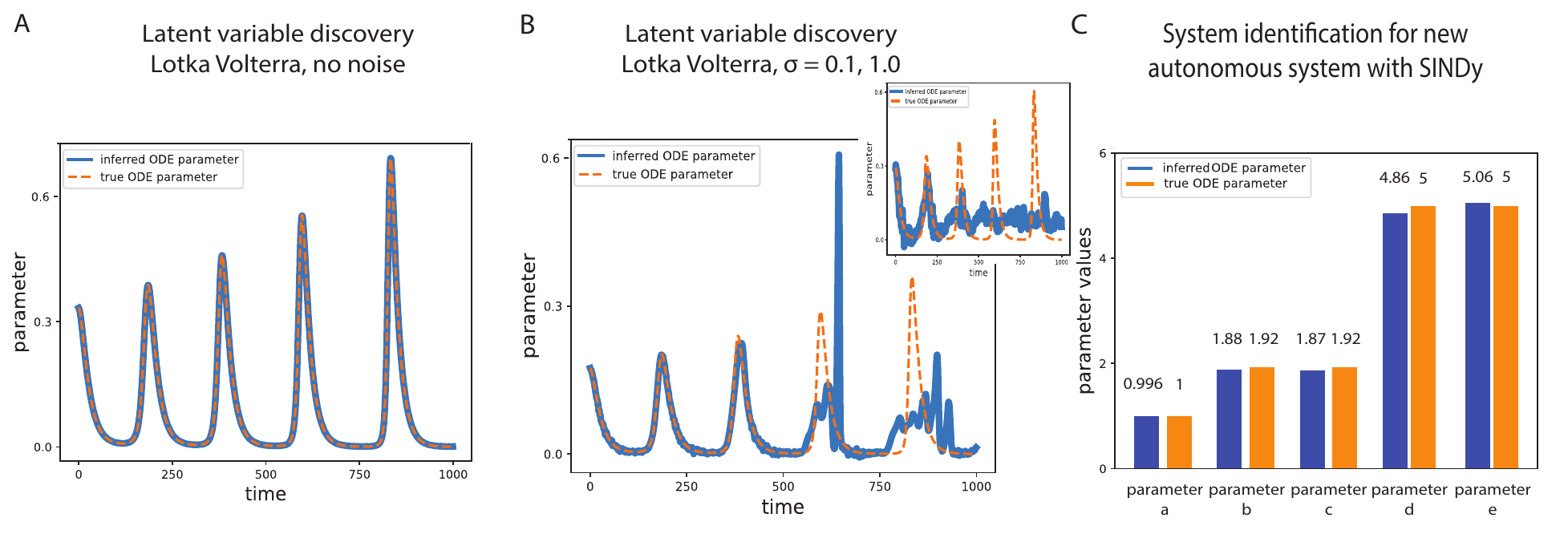}
\caption{Dynamic SINDy can be used for latent variable discovery. (A). Inferred (blue) versus true (orange) $y$ time series, from noiseless Lotka-Volterra. (B). Same as in (A), when noise of different standard deviations $\sigma$ is added to the Lotka-Volterra trajectory. (C). Inferred versus true coefficients in the Lotka-Volterra 2D ODE system, using SINDy for system identification.}
\label{fig5}
\end{figure}
\subsection{Dynamics in the nematode C. elegans during locomotion}

Modern neuroscientific data is noisy, nonlinear, and incomplete, with recordings from hundreds or thousands of neurons, yet many network features and neurons remain unmeasured. This makes it a challenging test for dynamic SINDy’s ability to model such complex systems. To demonstrate our method's potential, we use a dataset of C. elegans neural activity (Sec. 3.1, \cite{kato}). Unlike previous approaches that rely on probabilistic state space models \cite{slds_c_elegans} or hidden Markov models \cite{markov_ex1, markov_ex2, markov_ex3}, our method uncovers a global nonlinear switching model \cite{fiessler_c_elegans, morrison_c_elegans}. This model captures key features of the neural data: two stable fixed points representing forward and reversal behaviors, transitions between them, and variability in those transitions, reflecting real neural dynamics.
\\
\\
We first apply PCA to the data from one animal to obtain low-dimensional dynamics that cluster according to behavioral states (Figure \ref{fig5}A,B). Notably, only two dimensions are necessary to differentiate between forward, backward, and turning behaviors, although differentiating between various types of turns requires more dimensions. For a minimally complex model, we focus on the neural trajectory described by the dominant PC mode and its derivative. Our goal is to identify a nonlinear, parsimonious, and global model of the form:
\begin{align}
\dot{x} &= y \\
\dot{y} &= f(x,\beta) + u(t)
\end{align}
where $x$ is the data projected onto the first principal component, $f$ is an unknown function, $u(t)$ is a potential switching or control signal, and $\beta$ is a vector of parameters we would like to fit to our data.
\\
\\
We apply dynamic SINDy to minimize the error between the model's derivatives $(\dot{x}(t), \dot{y}(t))$ and the dominant PC derivatives from the data. Following sections 4.1 and 4.3, we identify the sparsity pattern of the SINDy coefficients, enforcing $\dot{x} = y$. The method highlights the terms $1, x, y, x^2, x^3$ for describing $\dot{y}$ and calculates their time-varying coefficients (SM Sec. 3.1, 3.2). To further simplify the model, we set coefficients for all variables to be constant, except the flexible term $u$ which we can also reduce to a time series of switches (Figure \ref{fig5}D, see SM Sec. 3.2.1) without meaningfully affecting global dynamics. Converting $u$ into a switching signal simplifies this term, helping to regularize the model and improve interpretability. This approach aligns with previous studies showing bistability and sudden transitions in behavior. 
\\
\\
Our approach identifies a cubic function for the differential equation model: $\ddot{x} = f(x, \beta) + u(t) =  0.002 \cdot x^3 + 0.0087 x^2 - 0.22 \cdot y + 0.05 \cdot x + u_i$, where $u_i$ alternates between $u_0 = -0.266$ and $u_1=0.044$ (see SM Sec. 3.1 for details). Each time $u$ switches, the cubic function shifts, altering the fixed point that the trajectory converges to. This switching signal $u$ enables the transitions between the two fixed points, which correspond to forward and reversal behaviors. Overall, the reconstructed data captures key features like fixed points and transitions (Figures \ref{fig5}E-F, \ref{fig5}G). By labeling the trajectory based on behaviors, we align the inferred dynamics with the training data (Figure \ref{fig5}H).
\\
\\
The reconstructions are accurate regardless of whether we use the processed switching term $u$ (Figure \ref{fig5} D) or the original time series $u$ (Figure \ref{fig5} C). However, $u$ alone does not adequately explain the data; removing other terms leads to poor fits or instability. By systematically eliminating different terms, we find that all are essential for capturing the dynamics. When we initialize the inferred ODE system from different starting points and use $u$ from training, the resulting dynamics qualitatively match the data. This suggests that our method effectively avoids overfitting. Unlike Morrison et. al., which relies on selecting a model based on human-labeled behavioral states, dynamic SINDy is fully data-driven and does not require labeled data to partition the phase space \cite{morrison_c_elegans}. Additionally, unlike Fieseler et. al., our model accommodates nonlinear dynamics with two stable fixed points \cite{fiessler_c_elegans}. A key advantage of our ODE model is the potential for biologically interpretable parameters (see \cite{morrison_c_elegans}. SM Sec. 3.3 offers a more comprehensive discussion of the benefits of our framework, comparing our global nonlinear switching model to previous studies. In summary, we have demonstrated that dynamic SINDy can do automatic data-driven model discovery, generating a nonlinear model with minimal input from the data scientist.
\begin{figure}
\centering
\includegraphics[width=\textwidth]{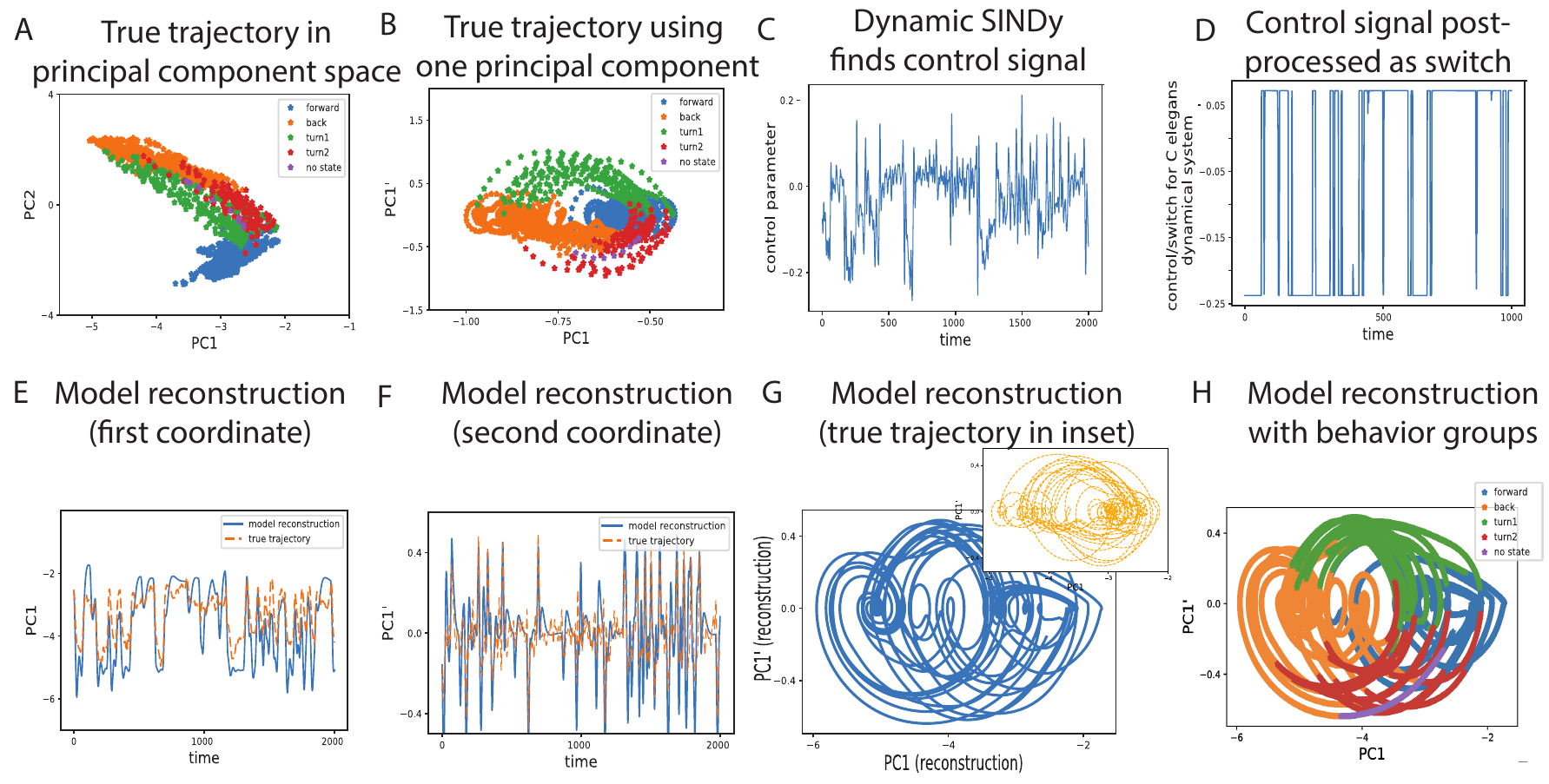}
\caption{(A). C. elegans neuronal activity is low-dimensional and clusters according to behavior; (B). Neuronal activity in phase space given by the first principal coordinate and its derivative; (C). Dynamic SINDy inferred constant term; (D). Processing coefficient in (C) as switch; (E) and (F) ODE model's match to ground truth trajectories; (G) (and (H)) 2D model trajectory (with labeled behavior).}
\label{fig5}
\end{figure}
\subsection{Dynamic SINDy and other methods for system identification}

\begin{figure}[h!]
\centering
\includegraphics[width=\textwidth]{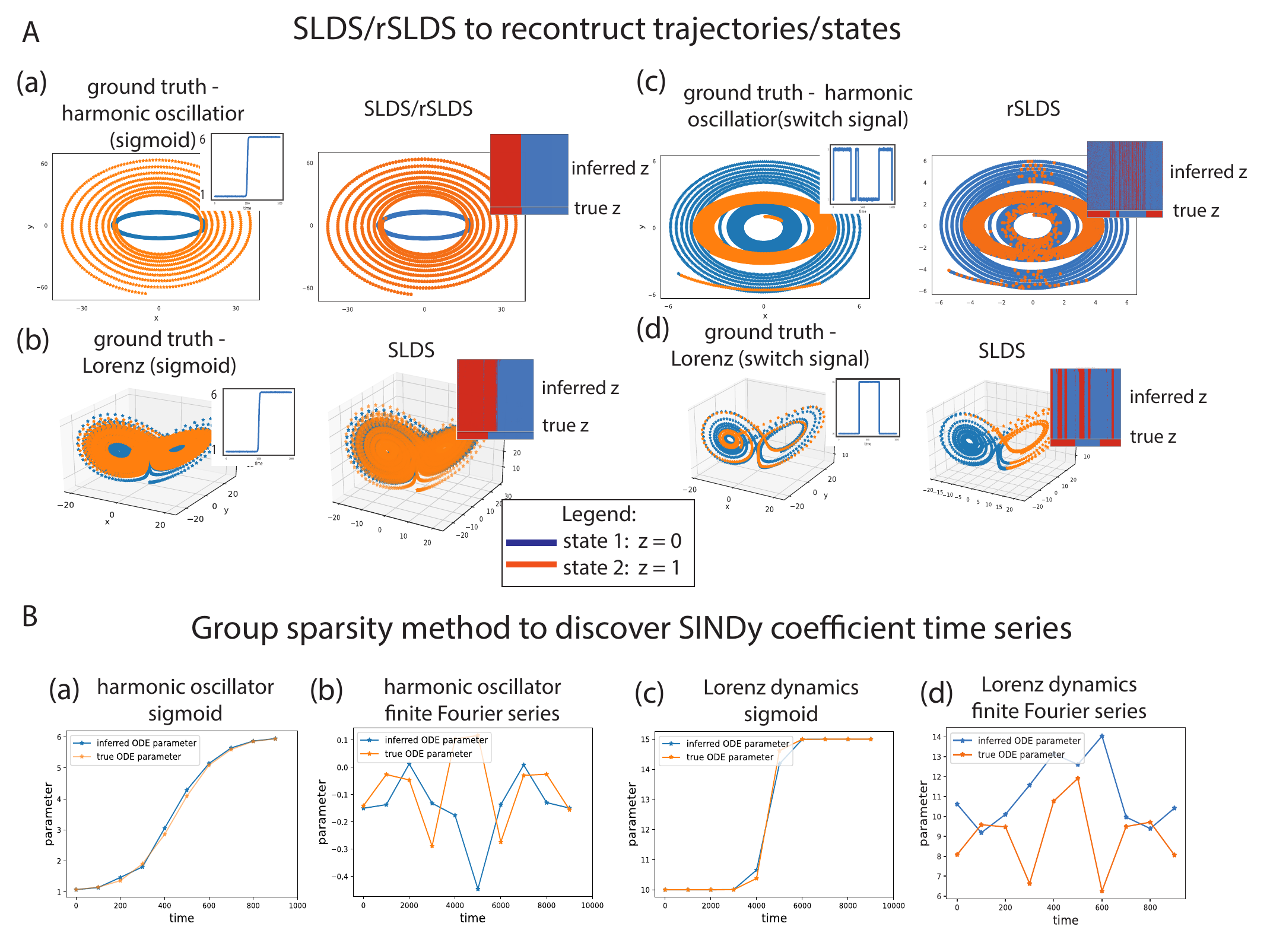}
\caption{(A) SLDS/rSLDS infers switching behavior for non-autonomous harmonic oscillators and Lorenz dynamics as coefficients vary. (a)-(d) left: ground truth dynamics, labeled switch values colored blue and orange. Inset shows true coefficient. (a)-(d) right: dynamics labeled by inferred switch. Inset: ground truth $z$ and samples of discrete latent values $z$ labeled by switch. (B) Inferred SINDy coefficients versus ground truth using group sparsity method.}
\label{fig_comparison}
\end{figure}

We begin by comparing dynamic SINDy with switching linear dynamic systems (SLDS) \cite{classical_SLDS1, classical_SLDS2} and its extension, rSLDS \cite{slds}. SLDS uses a discrete latent variable, $z_t$, to partition the state space between switches (see SM Sec. 4.1), simplifying complex nonlinear dynamics into more manageable linear segments. The rLDS extension allows switches to depend on continuous latent states and external inputs using logistic regression \cite{slds}. We evaluate how well SLDS/rSLDS identifies switching signals in the dynamical systems studied so far, specifically inferring where the latent variable $z$ changes for switching to occur. We use coefficients based on sigmoids and two types of switching signals (refer to the "ground truth" in Figure \ref{fig_comparison}A). Running SLDS or rLDS generates samples of the latent variable $z$ that segment the training trajectory, enabling us to compare this segmentation with the actual ground truth switches.
\\
\\
We find that for a sigmoidally varying coefficient, SLDS identifies the switch fairly well (Figure \ref{fig_comparison}A (a), (b)), as shown by the colored trajectories and the insets comparing the $z_t$ time series to the ground truth; although for Lorenz dynamics, the predicted change in the latent state $z_t$ is slightly delayed relative to the actual switch (Figure \ref{fig_comparison}A (b)). However, SLDS struggles when there are multiple state switches in the time series (Figures \ref{fig_comparison}A (c) and (d)). For the harmonic oscillator, rSLDS produces a model with too many switches and is more complex than the ground truth. For Lorenz dynamics, both SLDS and rSLDS switch periodically whenever the dynamics change between attractors, but this periodicity does not match the true switches defined by the coefficients. To address these challenges, we added time as a new dimension to the dataset, represented as a simple feature vector $[1,2,…,T]$, where $T$ is the total number of time steps. The goal was for SLDS/rSLDS to recognize that the switches are time-dependent rather than state-dependent. However, this addition did not improve the performance of SLDS or rSLDS.
\\
\\
Another method for identifying non-autonomous systems, discussed in references \cite{rudy_pde_group, group_sparsity_method} and SM Sec. 4.2, involves dividing the trajectory into smaller time windows and applying SINDy to each segment while enforcing a consistent sparsity pattern across all windows. We tested this approach on two toy datasets, using SINDy coefficients modeled as sigmoids, sinusoidal functions, and a Fourier series with seven frequencies. Without the group sparsity regularization, the sparsity patterns varied across the windows, highlighting the importance of group sparsity for achieving a coherent solution. The group sparsity approach worked well when the coefficients were sigmoid functions with varying smoothness (Figure \ref{fig_comparison} B (a) and (c)). However, it struggled with sinusoidal and Fourier series coefficients, particularly in the Lorenz system (\ref{fig_comparison} B(b) and (d)). In cases of misidentified coefficients, the algorithm also generated incorrect sparsity patterns. We conclude that neither SLDS or rSLDS, nor the group sparsity method are as effective as our method in identifying non-autonomous dynamical systems from data.

\section{Conclusion}
We have developed dynamic SINDy, an extension of SINDy designed for data-driven identification of noisy, nonlinear, and non-autonomous dynamical systems, as well as for discovering latent variables. 
We demonstrated the effectiveness of dynamic SINDy on both benchmark synthetic datasets and a real, noisy, chaotic dataset of neuronal activity from C. elegans. \\
\\
However, our method has some limitations. First, the DVAE architecture has many hyperparameters to tune, and results may not be robust to these settings, especially in noisy datasets. A systematic approach for hyperparameter tuning and addressing multiple solutions is necessary. To prevent overfitting, we should encourage simpler time series through regularization. Although we included a simple approximation of total variation term in our loss function, realistic datasets might require more sophisticated regularization. Future research should explore different DVAE architectures to evaluate their accuracy in reproducing dynamics and their ability to quantify uncertainty. It would also be valuable to apply our method to experiments with different trial dynamics and types of noise than those studied here (e.g., varying switching points across trials). Lastly, we aim to apply dynamic SINDy to realistic data from fields like biology, physics, and engineering to uncover hidden dynamics and fully utilize its potential for discovering latent variables.

\section{Supplemental information}
Supplemental Materials are included with the main manuscript. A repository with the code to generate the figures in the main paper can be found at \url{https://github.com/dvoina13/Dynamic_SINDy}.

\section{Acknowledgements}
The authors acknowledge support from the National Science Foundation AI Institute in Dynamic
Systems (grant number 2112085). We benefited from conversations with Preston Jiang, Ryan Raut,
and Sam Otto.
\\
\\
Funding sources have no direct involvement in the study design, in the analysis,
interpretation of results, or in the writing of the manuscript.

\bibliographystyle{plain}
\bibliography{arxiv_paper.bib}

\include{Supplementary}
\end{document}

%% file: Supplementary.tex
\title{Supplementary Material\\ Deep Generative Modeling for Identification of Noisy, Non-Stationary Dynamical Systems}

\setcounter{section}{0}
\setcounter{figure}{0}    

\maketitle
\section{Methods}

\subsection{Variational Autoencoders}

In this section, we elaborate on the mathematical foundation of the Variational Autoencoder (VAE) architecture \cite{vae_kingma, vae_rezende}.
\\
\\
Like standard autoencoders, VAEs have an encoder and decoder network to process input data and generate output. However, instead of mapping inputs to fixed points in the latent space, the encoder maps them to a probability distribution. The decoder then samples from this distribution to reconstruct the input. This probabilistic framework reduces overfitting by introducing variability into the latent space. After computing the reconstruction error, the network is trained via backpropagation, with the VAE relying on the reparameterization trick to ensure gradients can be propagated through the network.
\\
\\
Mathematically, we aim to approximate the data distribution $p^{*}(X)$ of some given observations $X$. When direct computation is intractable, we introduce a latent variable $z$ such that $p^{*}(x)$ is decomposed as:

\begin{align}
p^{*}(x) = \int_z p^{*}(x|z) p^{*}(z) dz 
\end{align}

where $p^{*}(x|z)$ is the likelihood and $p^{*}(z)$ is a prior, often set to a standard normal distribution. Since this integral is difficult to compute, we approximate $p_{\theta}(x|z)$ with a neural network parameterized by $\theta$. To estimate the posterior distribution $p^{*}(z|x)$, we approximate it with another neural network $q_{\phi}(z|x)$, parameterized by $\phi$. This is the core idea of variational inference: complex distributions are approximated by simpler, parametrized ones through optimization. We arrive at the following objective:

\begin{align}
\log p_{\theta}(x) &= \log \int_z p_{\theta}(x,z) dz \nonumber\\
&= \log \int_z p_{\theta}(x,z) \frac{q_{\phi}(z|x)}{q_{\phi}(z|x)} dz \nonumber \\
&= 
\log E_{z \sim q_{\phi}(z|x)} [\frac{p_{\theta}(x,z)}{q_{\phi}(z|x)}] \geq E_z [\frac{\log p_{\theta}(x,z)}{ q_{\phi}(z|x)}]
\end{align}

by Jensen's inequality. This leads to the evidence lower bound (ELBO):

\begin{align}
\mathcal{L} = E_z \frac{\log p_{\theta}(x,z)}{ q_{\phi}(z|x)}
\end{align}

The ELBO sets a lower bound for the evidence of observations and maximizing $\mathcal{L}$ will increase the log-likelihood of $X$. To find the parameters $\theta, \phi$ so as to maximize the ELBO, it is convenient to re-write $\mathcal{L}$ in the following way:
\begin{align}
\mathcal{L} &= \int_z q_{\phi}(z|x) \log(\frac{p_{\theta}(x,z)}{q_{\phi}(z|x)}) \nonumber \\
&= \int_z q_{\phi}(z|x) \log(\frac{p_{\theta}(x|z)p(z)}{q_{\phi}(z|x)}) = E_{z \sim q_{\phi}(z|x)} \log(p_{\theta}(x|z)) - D_{KL}(q_{\phi}(z|x) || p(z))
\end{align}

where $D_{KL}$ is the Kullback-Leibler divergence between the approximate posterior $q_{\phi}(z|x)$ and the prior $p(z)$. The ELBO comprises two terms: the expected log-likelihood of the data, and a regularization term that enforces similarity between the posterior and the prior.
\\
\\
In the autoencoder perspective, the encoder network maps inputs to the latent space via $q_{\phi}$, and the decoder maps the latent variables back to the input space via $p_{\theta}$. Both networks are trained jointly using stochastic gradient descent to optimize the ELBO. More details on the VAE framework can be found in Kingma et al.'s excellent review \cite{vae_kingma_review}.

\subsection{Dynamic VAEs}
Generating time series data presents unique challenges due to the intricate temporal relationships and the distribution of features at each time point. One common approach is using generative adversarial networks (GANs), which often incorporate recurrent neural networks (RNNs) for both generation and discrimination. However, despite numerous proposed architectures, GANs have struggled to capture the complex temporal dependencies inherent in time series data. Yoon et al. \cite{gen_gan} introduced a novel approach that blends the supervised training used in autoregressive models with the unsupervised training of GANs. While we experimented with this method for generating time series, the training proved to be time-consuming and impractical for our datasets (see Section 3.1). Further limitations are discussed by Desai et al. \cite{timeVAE}.
\\
As a result, we shifted our focus to methods based on Variational Autoencoders (VAEs) for time series, leveraging deep learning techniques to model complex temporal patterns more effectively.
\\
\\
An extensive review \cite{timeVAE_review} offers a unified framework for several VAE models extended to handle temporal and sequential data. These models, collectively referred to as dynamic VAEs (DVAEs), share common notation, methodology, and a standardized mathematical formalism. The review covers various approaches, including Deep Kalman Filters \cite{dvae11_dkf, dvae12_dkf}, Kalman Variational Autoencoders \cite{dvae2_kva}, Stochastic Recurrent Networks \cite{dvae3_srn}, Variational Recurrent Neural Networks \cite{dvae41_vrnn, dvae42_vrnn}, Stochastic Recurrent Neural Networks \cite{dvae5_srnn}, Recurrent Variational Autoencoders \cite{dvae6_rva}, and Disentangled Sequential Autoencoders \cite{dvae7_dsa}. In the following section, we will expand on the mathematical framework common to these methods, as outlined in \cite{timeVAE_review}.
\\
\\
Briefly, given a time-series $X_{1:T}$, and assuming latent variables $Z_{1:T}$, the goal is to specify the joint distribution of the observed and latent sequential data $p_{\theta}(X_{1:T}, Z_{1:T})$, where $\theta$ denotes the parameters of the true distribution's probabilistic model. DVAEs are hierarchical models in which both observed and latent variables are treated as time-ordered vectors. These models are often causal, meaning the distribution of variables at time $t$ depends only on previous time steps. This causality imposes the following factorization:
\begin{align}
p(X_{1:t}, Z_{1:T}) = \prod_{t=1}^{T} p(x_t, z_t | x_{1:t-1}, z_{1:t-1}) = \prod_{t=1}^{T} p(x_t | x_{1:t-1}, z_{1:t}) p(z_t | x_{1:t-1}, z_{1:t-1})
\label{joint_dvae}
\end{align}
The joint distribution of observed and latent variable sequences can be factorized using the chain rule. Crucially, different models proposed in the literature make different conditional assumptions to simplify the dependencies in the conditional distribution. For example, a simple model may make the following simplifications: $p(X_t | X_{1:t-1}, Z_{1:t}) = p(X_t | Z_t)$ and $p(Z_t | X_{1:t-1}, Z_{1:t-1}) = p(Z_t | Z_{t-1})$. In addition, different models may implement different network architectures to approximate $p_{\theta}$ and $q_{\phi}$. A detailed account of the kind of assumptions that each model implements to simplify 
(\ref{joint_dvae}) can be found in \cite{timeVAE_review}.

\subsubsection{timeVAE}

\begin{figure}[h!]
\centering
\includegraphics[width=\textwidth]{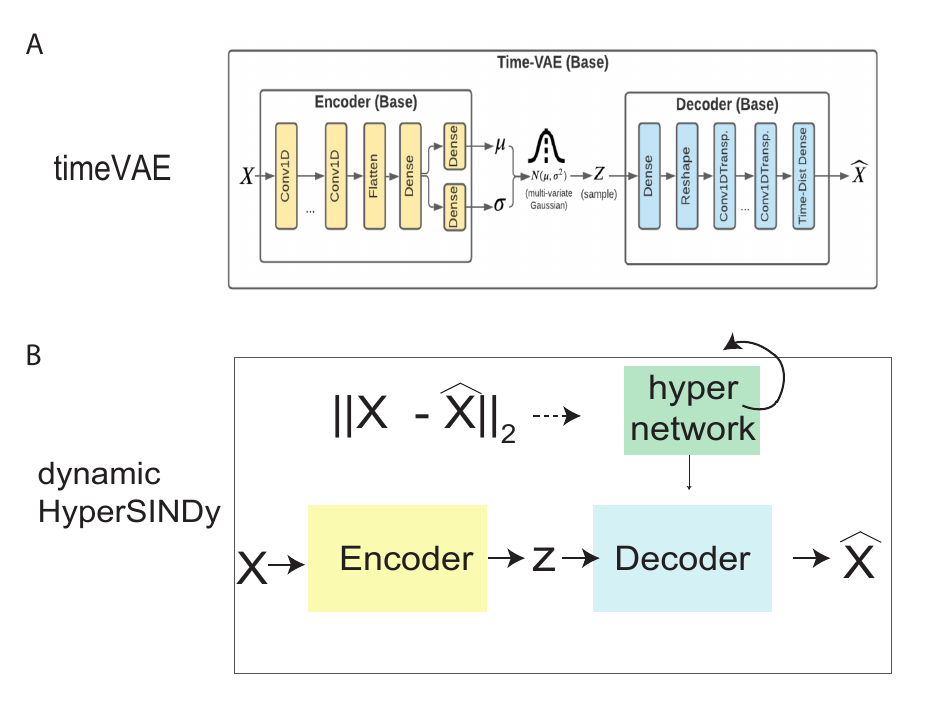}
\caption{(A) Schematic of timeVAE architecture from \cite{timeVAE}; (B) Schematic of dynamic HyperSINDy architecture described in SM Sec. 1.2.2}
\label{suppl_fig1}
\end{figure}

TimeVAE is a variational autoencoder designed to generate multivariate time-series data \cite{timeVAE}. It extends the standard VAE framework to model both the latent space and the temporal dependencies of a sequence of data vectors. Supplementary Figure \ref{suppl_fig1}A illustrates the basic TimeVAE architecture, which uses dense and convolutional layers without requiring specific time-series knowledge. The decoder allows for customizable distributions by adding layers to capture time-series components like level, trends, and seasonality, though we used the base version that excludes these custom structures in our experiments.
\\
\\
The input to the encoder is a 3D array of size $N \times D \times T$, where $N$ is the batch size, $D$ is the number of feature dimensions, and $T$ is the number of time steps. The encoder processes the data through convolutional layers with ReLU activations, flattens the output, and then applies a fully connected layer. The final encoder layer has $2d$ units, representing the mean and variance of a multivariate Gaussian distribution, where $d$ is the dimensionality of the latent space, a key hyperparameter. The reparametrization trick is used to sample from the Gaussian distribution, parameterized by the encoder's output.
\\
The decoder reconstructs the data by first passing the sampled latent vector $z$ through a fully connected layer, reshaping it into a 3D array, and processing it through a series of transposed convolutional layers with ReLU activation. The last time-distributed fully connected layer produces the final output that matches the original input dimensions.
\\
\\
Training TimeVAE involves optimizing the ELBO loss function (discussed in Section 1.1) with different weights on the reconstruction error and KL divergence between the approximate posterior $q_{\phi}(z|x)$ and the prior $p_{\theta}(z)$. Hyperparameters are tuned to determine the appropriate balance between reconstruction loss, KL divergence, and additional regularization terms (e.g., sparsity and total variation, for our problem set-up).
\\
\\
TimeVAE has been tested on four multivariate datasets \cite{timeVAE}: (1) a 5-dimensional sinusoidal dataset with varying frequencies, amplitudes, and phases; (2) a 6-dimensional stock market dataset from Yahoo Finance; (3) a 28-dimensional appliances energy prediction dataset from the UCI Machine Learning Repository; and (4) a dataset with 15 features of hourly air quality sensor readings from the UCI Machine Learning Repository. The results show that TimeVAE performs comparably to top generative models across various metrics, is computationally efficient, and outperforms existing methods in next-step prediction tasks, particularly when training data is limited \cite{timeVAE}.

\subsubsection{dynamic HyperSINDy}
Inspired by previous work \cite{preston_dpc, hypersindy}, we developed a hierarchical architecture to address non-autonomous systems, illustrated in Supplementary Figure \ref{suppl_fig1}B. The main text focuses on the dynamic SINDy framework with a timeVAE architecture, while Supplementary Figures \ref{suppl_fig3}, \ref{suppl_fig4}, \ref{suppl_fig5} show that incorporating dynamic HyperSINDy results in coefficients and trajectories that closely match the ground truth.
\\
\\
The first level consists of a standard VAE with encoder and decoder modules (SM Sec. 1.1). The decoder generates $X$ with a probability distribution $p_t(X)$ at each time step $t$. The next level introduces a hypernetwork, implemented as either a long short-term memory network (LSTM) or multi-layer perceptron (MLP), which updates the decoder’s weights to adjust the probability distribution for the following time step, allowing the system to capture temporal drift in the output:
\begin{align}
W_{\textrm{Decoder}}(t+1) &= W_{\textrm{Decoder}}(t) + \sum_{i=1}^{M} \alpha_i(t) \cdot D_i, \textrm{where    } \nonumber \\
\textrm{LSTM}(t) &=  \alpha(t) = [\alpha_1(t), ..., \alpha_n(t)]
\label{mix_basis}
\end{align}
Here, ${D_i}$ are fixed basis tensors to be learned, and $\alpha_i$ are hypernetwork outputs. This architecture adapts to changing dynamics, adjusting the decoder based on reconstruction error and updating the output probability distribution.
\\
\\

We modified this architecture for our problem. Instead of a VAE generating data, our decoder produces SINDy coefficients, which, when combined with the SINDy library, replicate system dynamics. The decoder approximates the true pdf of the SINDy coefficients. This setup builds on \cite{hypersindy} by adding a hypernetwork that updates decoder weights, forming what we call dynamic HyperSINDy. This extension allows for time-varying SINDy coefficients, processed sequentially rather than requiring the entire time series as input (as in timeVAE).
\\
\\
Two primary training methods are used for dynamic HyperSINDy:
\begin{itemize}
\item \textbf{Online Learning}: Ideal for switching systems where the network adapts as dynamics change. However, network parameters evolve, requiring tracking of parameter changes and identifying switch points after training. The hypernetwork is not needed in this setup.
\item \textbf{Alternate learning}: 
The hypernetwork is trained first with fixed main module parameters and basis tensors $D_i$, followed by adjustment of the main module parameters/$D_i$, while fixing the hypernetwork. This method is best for continuously varying SINDy coefficients, with LSTM as the preferred hypernetwork.
\end{itemize}

Training used the hyperparameters listed in Table \ref{hyperparams_dynamic_hypersindy}. We processed one trajectory at a time (trial batch size of 1) and used batch sizes of 1-10 time steps. The latent dimension of the VAE was set to 25 while the starting threshold was $0.1$. Every 50 epochs, we evaluated and set to zero any SINDy coefficients with a mean absolute value below a threshold. A relaxed L0 norm in the loss function encouraged sparsity in the SINDy coefficients, following \cite{hypersindy, L0_norm}.
\\
Several hyperparameters were increased progressively during training. The threshold rose by $0.005$ every $50$ epochs until it plateaued, alongside the weight $\lambda_{kl}$ for the KL divergence term, which increased until it reached a maximum value $\lambda_{max}$. The threshold plateaus as well once $\lambda_{max}$ is reached. We fixed the number of basis tensors $D_i$ to 10, which combined linearly with hypernetwork outputs to form decoder weights via Eq. (\ref{mix_basis}).
\\
\\
The encoder consisted of four fully connected layers with hidden dimensions of $64$, using ELU activation and an input dimension twice that of $X$, as it takes $X$ and $\dot{X}$ as input. The decoder also had four hidden layers, with a hidden dimension of $256$ and ELU activation. The hypernetwork, either an LSTM or MLP, contained two layers with an input dimension of $25$. We trained using the AdamW optimizer with an initial learning rate of $0.001$, weight decay of $1e-5$, gradient clipping at $1$, and Amsgrad. Additionally, an exponential learning rate scheduler with $\gamma = 0.999$ was used. Many hyperparameters match those in \cite{hypersindy}.

\begin{table}[h!]
\caption{Hyperparameters for Dynamic HyperSINDy}
\label{hyperparams_dynamic_hypersindy}
\begin{center}
\begin{tabular}{ll}
\multicolumn{1}{c}{\bf hyperparameter}  &\multicolumn{1}{c}{\bf value}
\\ \hline \\
 batch size (trials)        & 1 \\
 batch size (time steps)    & 1-10 \\
 latent variable dimension  & 25 \\
 threshold                  & 0.1 \\
 threshold interval         & 50 \\
 threshold increment        & 0.005 \\
 $\lambda_{kl}$             & 0.01 \\
 $\lambda_{kl}$ increment   & $\lambda_{kl}/5$\\
 $\lambda_{kl}$ max         & 1 \\
 M (number of basis tensors) & 10 \\
 hidden dim (decoder)       & 256 \\
 hidden dim (encoder)       & 64 \\
 input dim (LSTM)           & 25 \\
 gradient clip              & 1.0 \\
 cell dimension (LSTM).     & 30 \\ 
 optimizer                  & AdamW \\
 weight decay               & 1e-5 \\
 amsgrad                    & True \\
 learning rate              & 0.001 \\ 
 learning rate scheduler    & ExponentialLR \\
 gamma                      & 0.999
\end{tabular}
\end{center}
\end{table}

\subsection{Training dynamic SINDy with timeVAE: methodology and hyperparameters}

Training timeVAE requires normalizing the data beforehand. While \cite{timeVAE} normalizes by subtracting the minimum and dividing by the maximum to scale the data between 0 and 1, we normalize by dividing only by the maximum value. This normalization method affects the SINDy coefficients produced by our method, so we re-scale the resulting time-series before comparing them to the ground truth in synthetic datasets.
\\
\\
The loss function used to train our timeVAE architecture is:

\begin{align}
\textrm{loss} &= \lambda_{MSE} \cdot ||\hat{\dot{X}} - \dot{X}||_2^2 + \lambda_{KL} \cdot \textrm{KL div} + \lambda_{sp} \cdot <||\xi_{i,j}(t)||_1>_{i,j}  
\\
& + \lambda_{tv} \cdot \frac{< ||\xi_{i,j}(t+1) - \xi_{i,j}(t)||_1 >_{i,j,t}}{<||\xi_{i,j}(t)||>_{i,j,t} +\epsilon}
\label{eq_loss_}
\end{align}

where $\epsilon$ is the machine precision limit. The hyperparameters $\lambda_{MSE}, \lambda_{KL}, \lambda_{sp},$ and $\lambda_{tv}$ balance accuracy and complexity by adjusting the weights on the different loss terms: $\lambda_{MSE}$ controls the mean squared error, while the others handle regularization.
\begin{itemize}

\item The first term represents the mean squared error between the inferred derivative $\hat{\dot{X}}$ using dynamic SINDy and the derivative from the data $\dot{X}$. For all synthetic datasets, the ground truth derivative is the one used to obtain the trajectories $X$.

\item The second term is the Kullback-Leibler divergence (\textit{KL div}), a standard term in variational autoencoders (discussed in SM Sec. 1.1). It measures how closely the posterior distribution of $z$, as computed by the encoder given $X$, resembles the standard normal distribution. The KL divergence has an analytic form:

\begin{align}
\textrm{KL div} = -\frac{1}{2} \cdot  < (1 + 2\log(\sigma_{z_{i,j}}) - \sqrt{\mu_{z_{i,j}}} - \exp{(2\log{\sigma_{z_{i,j}}})}) >_{i,j}
\end{align}

where $\langle \cdot \rangle$ indicates averaging over latent dimensions $i$ and data points $j$, and $\mu_{z_j}$ and $\sigma_{z_j}$ represent the mean and standard deviation of $z_j$, with $\mu_{z_j}$ and $\log(\sigma_{z_j})$ as the encoder outputs for each input $X_j$.

\item The third term in Eq. (\ref{eq_loss_}) is a sparsity penalty that encourages some SINDy coefficients to be zero.

\item The fourth term is a normalized total variation penalty that prevents drastic changes in the solution over time. 

\end{itemize}

\subsubsection{Non-autonomous harmonic oscillators}

For the non-autonomous harmonic oscillators, we use the hyperparameters in Table \ref{hyperparams_timeVAE_harmonic_osc} to train the timeVAE architecture. These remain constant across datasets, despite differences in the time-varying coefficients $A(t)$ and $B(t)$. However, key hyperparameters like $\lambda_{sp}$ and $\lambda_{tv}$ vary depending on the dataset, as shown in Table \ref{hyperparams_timeVAE_harmonic_osc2}. Training is performed using the ADAM optimizer with a weight decay of $1e-5$ and gradient clipping at $1$.

\begin{table}[t]
\caption{Hyperparameters for timeVAE (non-autonomous harmonic oscillator)}
\label{hyperparams_timeVAE_harmonic_osc}
\begin{center}
\begin{tabular}{ll}
\multicolumn{1}{c}{\bf hyperparameter}  &\multicolumn{1}{c}{\bf value}
\\ \hline \\
 batch size     &       1  \\  
\hline
 latent dimension & 2  \\
 \hline

 threshold & 0.01 \\
 \hline
 
 library size & 3 \\
 \hline

 $\lambda_{MSE}$ & 3 \\
 \hline

 $\lambda_{KL}$ & 1000\\
 
 \hline
 \end{tabular}
\end{center}
\end{table}


\begin{table}[t]
\caption{Hyperparameters for timeVAE at different phases of training (non-autonomous harmonic oscillator)}
\label{hyperparams_timeVAE_harmonic_osc2}
\begin{center}
\begin{tabular}{llll}
\multicolumn{1}{c}{\bf hyperparameters}  &\multicolumn{1}{c}{\bf dataset}&\multicolumn{1}{c}{\bf at first training phase}&\multicolumn{1}{c}{\bf at second training phase}
\\ \hline \\
$\lambda_{sp}$ & A(t) sigmoid & 50  & 0 \\
\cline{2-4}
    & A(t) switch signal 1 & 500 & 0 \\\cline{2-4}
    & A(t) switch signal 2 & 200 & 0 \\\cline{2-4}
    & A(t) finite Fourier series & 1 & 0 \\\hline
$\lambda_{tv}$ & A(t) sigmoid & 100 & 1000\\\cline{2-4}
    & A(t) switch signal 1 & 100 & 1000 \\\cline{2-4}
    & A(t) switch signal 2 & 100 & 1000 \\\cline{2-4}
    & A(t) finite Fourier series 1 & 0 &2 \\\hline
 \end{tabular}
\end{center}
\end{table}

\subsubsection{Lorenz dynamics}
For the results in Sec. 4.3, involving the chaotic system with a time-varying parameter in the Lorenz dynamics, we follow the same steps as before (data normalization/post-processing, loss function, and two-stage training: first for sparsity pattern, then for coefficient recovery), but with different hyperparameters listed in Table \ref{hyperparams_timeVAE_lorenz}. These hyperparameters remain constant, regardless of how the Lorenz parameters vary over time.
\\
\\
During the first training stage, when the sparsity penalty is non-zero, the batch size is set to $10$ to ensure the correct sparsity pattern is learned. We use RMSProp with a weight decay of $10^{-5}$ and gradient clipping at $10$. The threshold gradually increases from $0.05$ to $0.1$ in increments of $0.025$ per epoch, while $\lambda_{sp}$ rises from $0$ to $20$ in steps of $1$. $\lambda_{tv}$ is fixed at $1000$. This gradual increase in hyperparameters follows a successful approach from a related study \cite{hypersindy}.

\begin{table}[t]
\caption{Hyperparameters for timeVAE (Lorenz dynamics)}
\label{hyperparams_timeVAE_lorenz}
\begin{center}
\begin{tabular}{ll}
\multicolumn{1}{c}{\bf hyperparameter}  &\multicolumn{1}{c}{\bf value}
\\ \hline \\
 latent dimension & 5  \\
 \hline

 library size & 3 \\
 \hline

 $\lambda_{MSE}$ & 3 \\
 \hline

 $\lambda_{KL}$ & 1000\\
 \hline
 \end{tabular}
\end{center}
\end{table}

\subsubsection{Lotka Volterra}
The incomplete Lotka Volterra system has only one variable $x$, therefore the library has three terms: $x,x^2,x^3$.  For training, we use the hyperparameters listed in Tables \ref{hyperparams_lotka_volterra} and \ref{hyperparams_lotka_volterra2}.

\begin{table}[h!]
\caption{Hyperparameters for timeVAE (Lotka Volterra)}
\label{hyperparams_lotka_volterra}
\begin{center}
\begin{tabular}{ll}
\multicolumn{1}{c}{\bf hyperparameter}  &\multicolumn{1}{c}{\bf value}
\\ \hline \\
  batch size & 1 \\
  \hline
 latent dimension & 2  \\
 \hline

 library size & 3 \\
 \hline

 $\lambda_{MSE}$ & 3 \\
 \hline

 $\lambda_{KL}$ & 1000\\
 \hline
 \end{tabular}
\end{center}
\end{table}

\begin{table}[h!]
\caption{Hyperparameters for timeVAE at different phases of training (Lotka Volterra)}
\label{hyperparams_lotka_volterra2}
\begin{center}
\begin{tabular}{lll}
\multicolumn{1}{c}{\bf hyperparameters}  &\multicolumn{1}{c}{\bf at first training phase}&\multicolumn{1}{c}{\bf at second training phase}
\\ \hline \\
 $\lambda_{sp}$ & 0.1 & 0 \\
 \hline
 $\lambda_{tv}$ & 0 & 0 \\
\hline
 \end{tabular}
\end{center}
\end{table}

\section{Latent variable discovery}

\subsection{Non-autonomous harmonic oscillator}
We can use the same approach with the non-autonomous harmonic oscillator as with the Lotka-Volterra system. We set $A(t) = -4$ and vary $B(t)$ sigmoidally such that $B(t) = 2 + \frac{1}{1 + exp(5 + t)}$. After dynamic SINDy identifies a trajectory for $B$, we add it to $(x,y)$ to form a 3D dynamical system. Using SINDy on $(x,y,B)$, we discover the following ODE which is almost exactly identical to the true dynamics, given that $B$ is a sigmoid that can be described by $\dot{B} = -6 + 5B - B^2$:
  \begin{align}
    \dot{x} &= -3.997 y \nonumber \\
    \dot{y} &= 1 \cdot B x \\
    \dot{B} &= -5.875 + 4.903 B -0.981 B^2 \nonumber
  \end{align}

\section{Dynamic SINDy for system identification of neuronal dynamics in the nematode C. elegans}

\subsection{Results}

Like in Morrison et. al., we have discovered a dynamical system model that switches between two stable fixed points. The differential equation model is expressed through a cubic function: $\dot{x} = a\cdot x^3 + b \cdot x^2 + c \cdot x + d \cdot y + u$ with distinct fixed points corresponding to the different switching states of $u$. 
More precisely, the differential model inferred has the form:
\begin{align}
    &\dot{x} = y \nonumber \\
    &\dot{y} = -0.002 \cdot x^3 + 0.0087 x^2 - 0.22 \cdot y + 0.05 \cdot x + u_i, \hspace{0.1in} i=1,2 \\
     & u_0  \approx -0.266; u_1 \approx 0.044. \nonumber
     \label{final_Celegans_model}
\end{align}

When $u = u_0 < 0$, the dynamical system has one stable fixed point at $-5.25$ (the other roots of the cubic equation are complex). This fixed point corresponds to the reversal behavior. Then, when $u = u_1 >0$, the dynamical system has two stable and one unstable fixed point: $-2.32$ and $7.88$ stable fixed points and $-1.19$ unstable. Therefore, varying $u$ can generate a bifurcation. In practice, the trajectory shifts between $-5.25$ and $-2.32$ in tandem with behavioral switches between reversal and forward states.
\\
\\
The two-dimensional model is a simple model that fits the first principal component and captures stable state clusters and turning trajectory variability. Once the low-dimensional coordinates are identified, dynamic SINDy effectively enables data-driven model discovery. Future work will extend this approach to multiple animals to test its generality across individuals.

\subsection{Training dynamic SINDy on C. elegans data}

We apply timeVAE to infer dynamics for a single worm as a proof-of-concept, demonstrating dynamic SINDy's capability for data-driven discovery. Since there is only one trajectory per worm, uncertainty quantification isn't possible. As with synthetic datasets, we normalize the trajectories and train timeVAE to infer the differential equation's sparsity pattern using the hyperparameters from Tables \ref{hyperparams_celegans1} and \ref{hyperparams_celegans2}.
\\
\\
With a threshold of $0.01$, only the terms $1$, $x$, and $x^2$ are considered important. We add $y$ and $x^3$ for comparison with the Morrison et al. model and retrain with a fixed sparsity pattern, omitting sparsity regularization from the loss. Our SINDy coefficients vary over time, matching behavioral transitions between forward and reversal locomotion. Increasing total variance regularization was not effective, so we averaged non-constant SINDy coefficients over time to simplify the model. This is part of the training process, where we take the average of all non-constant coefficients at the last layer of the network to yield the model output and backpropagate. The constant SINDy coefficient is not constrained, but all other coefficients do not change in time. 
\\
\\
The resulting differential equation model, detailed in Eq. \ref{final_Celegans_model}, includes the time series $u(t)$, which is shown in Figure 6C (main text). We interpret $u(t)$ as a switching variable and hypothesize that even a simple switching time series can qualitatively capture the neural activity data. To test this hypothesis, we post-process the $u(t)$ time series to generate a switch-like signal.

\subsubsection{Post-processing the switching signal}
Starting with the $u(t)$ time series inferred using dynamic SINDy, we perform the following steps:

\begin{itemize}
\item Subtract the mean of $u(t)$ over time. We also note the approximate minimum and maximum values, which will be used later. 

\item Scale the data by a large factor (1000) and apply a pointwise sigmoid function across time, producing a time series of switches between 0 and 1.

\item Finally, re-scale the time series to vary between the previously determined minimum and maximum values, and then add back the mean $<u(t)>_t$ to obtain the final post-processed switching time series.
\end{itemize}

To evaluate the accuracy of our model, we integrate the differential equation from Eq. \ref{final_Celegans_model} using the post-processed switching signal $u(t)$ and compare the resulting trajectory to the real trajectory (Figure 6E-G, main text). Since the C. elegans data has low time resolution ($\Delta t = 0.35749752$), we interpolate the data using the CubicSpline function from the scipy.interpolate library. We reduce the time step to $\Delta t/100$ and perform numerical integration using the Euler method.

\begin{table}[t]
\caption{Hyperparameters for timeVAE (C. elegans data)}
\label{hyperparams_celegans1}
\begin{center}
\begin{tabular}{ll}
\multicolumn{1}{c}{\bf hyperparameter}  &\multicolumn{1}{c}{\bf value}
\\ \hline \\
 batch size     &       1  \\  
\hline
 latent dimension & 2  \\
 \hline

 threshold & 0.01 \\
 \hline
 
 library size & 3 \\
 \hline

 $\lambda_{MSE}$ & 3 \\
 \hline

 $\lambda_{KL}$ & 1000\\
 \hline
\end{tabular}
\end{center}
\end{table}

\begin{table}[t]
\caption{Hyperparameters for timeVAE at different phases of training (C. elegans data)}
\label{hyperparams_celegans2}
\begin{center}
\begin{tabular}{lll}
\multicolumn{1}{c}{\bf hyperparameters}  &\multicolumn{1}{c}{\bf at first training phase}&\multicolumn{1}{c}{\bf at second training phase}
\\ \hline \\
$\lambda_{sp}$  & 10  & 0 \\
$\lambda_{tv}$  & 100 & 1000\\
\hline
 \end{tabular}
\end{center}
\end{table}

\subsection{Background: related studies and comparisons to our model}
\subsubsection{Comparison with state space models}
Our findings with dynamic SINDy reveal a key similarity with the probabilistic state space model proposed by Linderman et al. \cite{slds_c_elegans}: both models switch between different dynamical regimes. However, despite being nonlinear, our model is more parsimonious in several ways.
\\
\\
Linderman et al. propose a hierarchical recurrent state space model that switches between simple linear models, using Bayesian inference to fit the model at scale \cite{slds_c_elegans}. This model decomposes complex nonlinear neural activity into discrete states with simple linear dynamics, which correspond to behaviorally relevant aspects of the worm's behavior. The transition probabilities depend on both the preceding state and the position in continuous state space, with each discrete state largely tied to the activation of specific neuron clusters.
\\
\\
While this model provides insights into C. elegans neural dynamics, its linear state space models are local. The model switches between eight discrete states, each representing a smaller linear system fitted to the data to explain local dynamics \cite{slds_c_elegans}. In contrast, dynamic SINDy discovers a global nonlinear ODE model that switches between only two states. Thus, we simplify the model by replacing eight local linear regimes with a more compact nonlinear system switching between two states.
\\
\\
A future direction is to develop a generative model for the switching behavior of $u(t)$, possibly using a probabilistic model or differential equation linked to the variables $x$ and $y$. This would allow us to eliminate the dynamic SINDy network post-training, retaining a global nonlinear switching differential equation with just four parameters, compared to the many more parameters required by the hierarchical recurrent SLDS for its eight linear systems (even when considering only the continuous variable dynamics for a fair comparison with our approach in PC space).
\\
\\
Moreover, the hierarchical recurrent SLDS is a statistical model that doesn’t directly map onto network dynamics or account for biologically realistic state transitions. While further research is needed to validate our model’s connection to biological measurements of neural activity, nonlinear differential equations like ours are potentially more interpretable. For example, a single parameter change in a global nonlinear model similar to ours can reproduce different long-timescale behaviors observed in C. elegans \cite{morrison_c_elegans} (see Sec. 3.1.4 below). This modulation mirrors distinct changes in state distribution and switching frequencies seen in experiments, which are linked to specific neuromodulators and neurons \cite{c_elegans_experiment1, c_elegans_experiment2}.
\\
The challenges discussed here also apply to simpler models based on Markov dynamics, such as hidden Markov models (HMMs) \cite{markov_ex1, markov_ex2, markov_ex3}.

\subsubsection{Comparison with a nonlinear global model with control}
Our C. elegans neural activity modeling is inspired by Fieseler et al. and Morrison et al. \cite{fiessler_c_elegans, morrison_c_elegans}. Unlike state space models, Morrison et al. discovered a minimally parameterized global nonlinear model with control that mimics Hidden Markov model state transitions within a single dynamical system. This model captures key features of the C. elegans calcium imaging data, including two stable fixed points for forward and reversal behaviors, state transitions triggered by control signals, and variability in transition trajectories that match neural activity data \cite{morrison_c_elegans}.
\\
\\
The model is represented as:
\begin{align}
\dot{\textbf{x}} = F(\textbf{x},\beta) + \textbf{u}(t) 
\end{align}
where $\beta$ governs longer timescale dynamics, and $u(t)$ is a control signal operating on faster timescales. $u(t)$ is a one-dimensional signal that may integrate multiple local and non-local processes. This separation of intrinsic dynamics and control inputs increases the model's interpretability.
\\
\\
Nonlinear control has been used in other biological networks to describe switching between multiple stable states \cite{nonlinear_control1, nonlinear_control2, nonlinear_control3} A significant advantage is that a nonlinear model can have multiple fixed points corresponding to different behavioral states -- in the case of C. elegans, forward and reversal motion. A heuristic model capturing C. elegans behavior is:
\begin{align}
\dot{\textbf{x}} &= \textbf{y} \nonumber \\
\dot{\textbf{y}} &= -(\textbf{x}-1)(\textbf{x}-\beta)(\textbf{x}+1) + \lambda \textbf{y} + \textbf{u}(t)
\end{align}

When $u=0$ this cubic system has two stable fixed points at  $x = \pm1$ corresponding to forward and reversal motions, as well as an unstable point at $\beta$. The fixed points correspond to locations in the state space where $F(x, \beta) = 0$ and $u(t) = 0$. Transitions to the other stable fixed point occur when $u(t) \neq 0$, corresponding to a shift in the behavior of the animal. To capture the stochasticity of the data, additional noise terms are added to $\dot{x}, \dot{y}$.
\\
\\
The model is fit to reproduce the dominant PCA mode of the neural activity data. Importantly, the optimization is done by using the manually annotated behavioral labels to determine when the control switches values. For forward and reversal motion, $u=0$, while each type of turn (reversal to forward and vice versa) corresponds to a different $u$ value. These distinct models are fitted to the corresponding time series segments based on the annotations. The resulting nonlinear control model and the parameters found through optimization is fully described in \cite{morrison_c_elegans}.
\\
\\
Both our model and the global nonlinear model with control described above employ nonlinear terms in the dynamics. These are global models with few parameters that capture the most important qualitative features in the C. elegans data.\\
A key difference is that the control variable in Morrison et al. takes $3$ values, including $0$ during stable states, while dynamic SINDy’s switching variable takes $2$ values that influence the fixed points and are longer-lasting than the transient controls.
\\
\\
A key advantage of our method is that it is entirely data-driven, requiring no behavioral annotations or manual fitting. We directly input the low-dimensional neural activity time series, allowing dynamic SINDy to automatically discover the governing equations. This reduces the effort required from the data scientist while still capturing the system's essential dynamics.

\subsubsection{Comparison with a linear model with control}
A related study proposed a global linear model with control
whereby a linear dynamical system is actuated by temporally sparse control signals \cite{fiessler_c_elegans}. Denoting $\textbf{x}_j = \textbf{x}(t_j)$, neural activity across neurons at time $t_j$, and $\textbf{X}$ a matrix of neuronal data at different snapshots in time, $\textbf{X} = [\textbf{x}_1, ..., \textbf{x}_m]$, dynamic mode decomposition (DMD) provides a linear model for the dynamics of the state space:
\begin{align}
\textbf{X}' = \textbf{A}\textbf{X}
\end{align}
where $\textbf{X}'= [\textbf{x}_2, \textbf{x}_3, ... \textbf{x}_{m+1}]$ is offset by one time step compared to $\textbf{X}$. Since a linear model alone cannot capture the neural activity data, DMD with control (DMDc, \cite{dmd_c}) is employed to distinguish between the underlying dynamics and control signals $\textbf{U} =[\textbf{u}_1, \textbf{u}_2, … \textbf{u}_m]$, where $\textbf{u}_j = \textbf{u}(t_j)$ are actuation signals at a snapshot in time. DMDc regresses to the linear control system: 
\begin{align}
\textbf{X}’ = \textbf{A}\textbf{X} + \textbf{B}\textbf{U}
\end{align}

The control signal can either be fixed using manually annotated behavioral onsets in a supervised setting or learned jointly with $A$ and $B$ (Algorithm 1, \cite{fiessler_c_elegans}). To avoid trivial solutions, the control signals are constrained to be sparse, meaning transitions between states should be infrequent. The following loss function, incorporating an $l_0$ regularization, is minimized using the sequential least squares thresholding algorithm:
\begin{align}
\textrm{loss} = \min_{\textbf{A},\textbf{B},\textbf{U}} = ||\textbf{A}\textbf{X} + \textbf{B}\textbf{U} - \textbf{X}' ||_2 + \lambda ||\textbf{U}||_0
\end{align}
If control signals are internally generated, they are either random or encoded within the network. Sparse variable selection and time-delay embeddings test the influence of present and past data to determine which neurons predict the controls:
\begin{align}
\textbf{u}_k = \textbf{K}_1 \textbf{x}_k + \textbf{K}_2 \textbf{x}_{k-1} + … 
\end{align}
Key findings from this study include that the unsupervised algorithm produces control signals somewhat correlating with manually annotated behavioral onsets, though it fails to capture forward motion onsets. This suggests that neurons involved in forward motion \footnote{This comparison between model and data at the single neuron level is only possible due to identification of neurons with stereotyped identities, as described in \cite{kato}} require non-trivial nonlinearities throughout the time series for full reconstruction, not just control signals at the onset. Reversal neurons are well-modeled by the supervised control signals, implying fewer required nonlinearities other than the transition signal itself. Turns are also largely captured by the neuronal activity in specific cells, although there is much more variability.
\\
\\ 
The global linear framework, with internally generated control signals, partially explains neuronal activity but produces a weak qualitative and quantitative match with the data. For instance, the correlation between real neuronal activity and the model is at or only slightly above $0.5$, even for neurons not implicated in forward motion, that the model supposedly is successful in capturing. Incorporating manual annotations significantly improves the model's accuracy, but it shifts the approach from data-driven to supervised.
\\
\\
The inability to capture forward motion and the ineffectiveness of forward control signals demonstrate that nonlinearities are necessary for many interactions in the system. A linear model with control can produce only one fixed point, with all other states simply being longer-lived, conflicting with empirical evidence that both forward and reversal behaviors in C. elegans are stable states \cite{fiessler_c_elegans}. Empirical studies have instead shown that multiple behavioral states appear to be stable \cite{morrison_c_elegans}. 
\\
\\
These results support our approach using dynamic SINDy, which automatically identifies nonlinear systems with multiple fixed points, without requiring manual annotations. Dynamic SINDy discovers a dynamical system with switching signals as opposed to controls, while incorporating nonlinearities that could be critical in capturing complex neural dynamics. Furthermore, our framework, as described in Sec. 4.4, could discover how transition signals depend on the data to create a fully closed-loop feedback system. Future work should confirm that our low-dimensional model can explain neural activity according to cell class. 

\section{Other methods for system identification of non-autonomous dynamic systems}
\subsection{Switching Linear Dynamical System (SLDS)}

The generative model for switching linear dynamical systems (SLDS) is as follows: for each time $t = 1,2,...,T$, there is a discrete latent state $z_t \in \{1,2,..., K\}$ that follows Markov dynamics: $$z_{t+1} | z_t, \{\pi_k\}_{k=1}^K \sim \pi_{z_t},$$
where $\{\pi_k\}_{k=1}^K$ is the Markov transition matrix and $\pi_k \in [0,1]^K$ is the k-th matrix row. In addition to $z_t$, there is a continuous latent state $x_t \in \mathbb{R}^M$ following linear dynamics that depend on $z_t$:

\begin{align}
x_{t+1} = A_{z_{t+1}} x_t + b_{z_{t+1}} + v_t, v_t \sim N(0,Q_{z_t})
\end{align}

where $A_{z_t}, Q_{z_t}$ are matrices and $b_{z_t}$ is a vector depending on the latent state $z_k \in {1,...,K}$. In addition, we have access to observables $y_t$, generated from the continuous latent state $x_t$:
\begin{align}
y_t = Cx_t + d + w_t, w_t \sim N(0,S)
\end{align}
where $C, S, d$ are shared matrices and a vector across different discrete states $z_t$. We denote the complete set of parameters as $\theta = \{ \pi_k, A_k, Q_k, b_k, C, S, d | k = 1,...,K\}$ and learn SLDS using Bayesian inference and a set of convenient priors as detailed in \cite{slds}.
\\
\\
An extension of SLDS -- rsLDS -- allows the discrete switches to depend on the continuous latent state and exogeneous inputs through a logistic regression \cite{slds}. Specifically, when a discrete switch occurs whenever a continuous state enters a particular region of state space, SLDS is unable to learn this dependence, while rSLDS designed to address this state dependence. An important contribution of \cite{slds} is an inference algorithm leveraging Polya-gamma auxiliary variable methods to make inference fast, scalable, and, easy. We make use of this algorithm through the open-source rSLDS libraries \cite{code_slds}.

\subsubsection{Training}
To train the SLDS/rSLDS models, we first select the number of latent states in advance. For switch-like underlying coefficients, we choose two states to maximize SLDS's ability to identify the switching dynamics. We then perform principal component analysis (PCA) on the data, followed by fitting a simpler autoregressive hidden Markov model (AR-HMM), which lacks continuous latent states. The discrete latent variables $z$ inferred from the AR-HMM, along with the data projected onto a small number of principal components, are used to initialize the SLDS/rSLDS algorithms. The SLDS/rSLDS training algorithms are implemented from the following github repository: \url{https://github.com/slinderman/recurrent-slds}.

\subsection{A method based on group sparsity}
We focus on two studies \cite{rudy_pde_group, group_sparsity_method} that address system identification in non-autonomous systems. Both studies propose a method for identifying ODEs with time-varying SINDy coefficients by dividing the trajectory into smaller time windows and applying SINDy to each segment while maintaining the same sparsity pattern across all segments. In \cite{rudy_pde_group}, this sparsity pattern is enforced using group sparsity regularization, while \cite{group_sparsity_method} introduces a novel algorithm based on sequential thresholding least squares (STLQ) from the original SINDy paper \cite{brunton_sindy}. This STLQ adaptation averages the SINDy coefficients across time windows and compares the average to a threshold, setting coefficients below this threshold to zero.
\\
\\
The group sparse penalized method for model selection and parameter estimation is used with datasets of multiple trajectories that share the same physical laws, but differ in bifurcation parameters \cite{group_sparsity_method}. This framework is subsequently adapted to switching systems, whereby in a Lorenz system, the parameter $\alpha$ changes from $-1$ to $6.66$ at some unknown time. The framework matches our problem, therefore we adapt the algorithm proposed in this analysis to the non-autonomous dynamical systems we study (Sec. 3.3).
\\
\\
Adapting the notation in \cite{group_sparsity_method} to our own, we have a total of $M$ time windows that partition the trajectory, and we denote  time windows by $i$. Data points from specific time windows are indexed by superscript $i$, while different variables of the system are denoted by subscript $j$, $j=1, ..., n$. For instance, in the Lorenz system variables $x,y,z$ correspond to $x_{j}$, $j=1,2,3$, where $x_j^i$ corresponds to variable $x_j$ within time window $i$. We can then define variable $\Xi_j$ in terms of $\xi_j^{i}$ which are SINDy coefficients corresponding to variable $j$ within time window $i$:

$\Xi_j =
\begin{bmatrix}
 |        &   |         &      &    |    \\ 
\xi_j^{1} &  \xi_j^{2}  &  ... & \xi_j^{M}\\
 |        &   |         &        &    | 
\end{bmatrix} 	
$
\\
\\
Next we can define the data matrix $X^{(i)}$, the velocity matrix $\dot{X}^{(i)}$, and the dictionary matrix $\Theta^{(i)}$ as:
$X^{i} =
\begin{bmatrix}
 |      &   |        &      &    |    \\ 
x_1^{i} &   x_2^{i}  &  ... & x_n^{i} \\
 |      &   |        &      &    | 

\end{bmatrix} = \begin{bmatrix}
 x_1(t_1 ;\lambda^{(i)}) & x_2(t_1 ;\lambda^{(i)}) & ... & x_n(t_1;\lambda^{(i)})\\
 x_1(t_2 ;\lambda^{(i)}) & x_2(t_2 ;\lambda^{(i)}) & ... & x_n(t_2;\lambda^{(i)})\\
   ...      &    ...  & ...   & ... \\
x_1(t_{l_i} ;\lambda^{(i)}) & x_2(t_{l_i} ;\lambda^{(i)}) & ... & x_n(t_{l_i};\lambda^{(i)})
\end{bmatrix}
$
\\
\\
\\
\\
$\dot{X}^{i} = 
\begin{bmatrix}
 |        &   |         &      &    |    \\ 
\dot{x}_1^{i} &  \dot{x}_2^{i}  &  ... & \dot{x}_n^{i}\\
 |        &   |         &        &    | 

\end{bmatrix} = \begin{bmatrix}
 \dot{x}_1(t_1 ;\lambda^{(i)}) & \dot{x}_2(t_1 ;\lambda^{(i)}) & ... & \dot{x}_n(t_1;\lambda^{(i)})
 \\
 \dot{x}_1(t_2 ;\lambda^{(i)}) & \dot{x}_2(t_2 ;\lambda^{(i)}) & ... & \dot{x}_n(t_2;\lambda^{(i)})
 \\
   ...      &    ...  & ...   & ... \\
\dot{x}_1(t_{l_i} ;\lambda^{(i)}) & \dot{x}_2(t_{l_i} ;\lambda^{(i)}) & ... & \dot{x}_n(t_{l_i};\lambda^{(i)})
\end{bmatrix}
$
\\
\\
and
\\
\\
$\Theta^{i} = 
\begin{bmatrix}

\textbf{1}_{l_i,1} & X^{i} & (X^{i})^2 & (X^{i})^3 & ...

\end{bmatrix}
$

Using this notation, the optimization problem can be rewritten as a least-square fitting :
\begin{align}
\min_{\Xi_j} \sum_{i=1}^n ||\Theta^i \xi_j^i - \dot{X}_j^i ||_2^2
\end{align}
for each $j = 1,...,n$.

To help prevent overfitting, we add regularization to this cost function, by including a penalty on the number of active candidate functions. The main assumption of this method is that coefficients $\xi^{i}$ have the same support set (in $j$) for each $i$, but can differ in value. Thus we can group each row $j$ together to be either zero or nonzero, therefore the number of active (nonzero) rows in $\Xi_j$ is sparse. The cost function now implements the following group-sparse optimization problem:
\begin{align}
\min_{\Xi_j} \sum_{i=1}^m ||\Theta^{i} \xi_j^i - \dot{X}_j^i||_2^2 + \lambda || \Xi_j||_{2,0}
\end{align}
where the $l_{2,0}$ penalty is defined as:

\begin{align}
||A||_{2,0} = \#\{k: (\sum_{l=1}^{m} |a_{kl}|^2)^{1/2} \neq 0\}
\end{align}

for any matrix $A = [a_{k,l}]$. Although the problem is non-convex, the authors in \cite{group_sparsity_method} propose to solve it numerically using an iterative hard thresholding algorithm, the \textit{group hard-iterative thresholding algorithm for dynamical systems}:
\\
\\
Group Hard-Iterative Thresholding Algorithm for Dynamical Systems:
\begin{algorithmic}[1]
\label{group_sparsity1}

\State Given: initialization matrix $\Xi^{(0)}$, tol and parameters $\gamma$.\\
\While {$||\Xi^{(k+1)} - \Xi^{(k)}|| > tol$}  \\
   \For {i= 1 to m}\\
        \State $(\xi^i)^{(k+1)} = (\xi^i)^{(k)} - (\Theta^i)^T \left( \Theta^i (\xi^i)^{(k)} - \Xi^i \right) $\\
    \EndFor \\
    \State $S^{(k+1)} = supp(H_{\sqrt{\gamma}}[\xi^1, \xi^2, ..., \xi^m])$\\
    \For {i= 1 to m} \\
        \State $(\xi^i)^{(k+1)} = \textrm{argmin}_{\xi^i} ||\Theta^i \xi^i - \dot{X}^i ||_2^2$ s.t. $supp(\xi^i) \subset S^{k+1}$\\
    \EndFor\\
\EndWhile \\
\end{algorithmic}

where $supp(x)$ is the support set of $x$, i.e. the indices of $x$ that correspond to the nonzero values.
\\
\\
While we implement this algorithm and test it on the data, we have found, surprisingly, that a simpler algorithm is more effective in many cases:
\\
\\
Simple sequential thresholding algorithm:
\begin{algorithmic}
\label{group_sparsity2}

\State Solve $\Theta^i \Xi^i = \dot{X}^{i}$ for each time window $i$ and trajectory $X$ and stack these least squares results in a matrix $\widetilde{\Xi}^{(0)}$ to be used as initial condition
\State Choose threshold
\For {j = 1 to 100} \%100 iterations
    \State \%average over time windows \textit{s} and compare to threshold:
    \State $\textrm{smallinds}  \gets \{ k_1, k_2 \hspace{0.1in} |  \hspace{0.1in} < |\widetilde{\Xi}^{(j-1)}[k_1, k_2,s]| >_s  <  \textrm{threshold}\}$ 
    \State $\textrm{biginds}  \gets \{ k_1, k_2 \hspace{0.1in} |  \hspace{0.1in} > |\widetilde{\Xi}^{(j-1)}[k_1, k_2,s]| >_s  >  \textrm{threshold}\}$

    \For {i = 1 to m}
        \State Solve for $\Xi^i$: $\Theta^i[\textrm{biginds}]  \cdot (\Xi^i[\textrm{biginds}])^{(j)} = \dot{X}^{i}$
        \State $(\Xi^i[\textrm{smallinds}])^{(j)} = 0$ \%coefficients are 0 all across time windows i

    \EndFor
     \State $\widetilde{\Xi}^{(j)} =  [(\Xi^1)^{(j)}, ..., (\Xi^m)^{(j)}]$
\EndFor
\end{algorithmic}

Comments starting with $``\%"$ are provided throughout the code to clarify its meaning. We use this algorithm throughout to showcase our results using the group sparsity method.
\\
\\
For training, we have varied the total time for the trajectories, the number of batches used, the time window, as well as the precise algorithm used. Throughout these experiments we have found that the algorithm was highly sensitive to whether the data was normalized or not, specifically we have found that \textbf{not} normalizing the data yielded superior results.


\newpage

\title{Supplementary Figures}
\maketitle

\begin{figure}[h!]
\centering
\includegraphics[width=\textwidth]{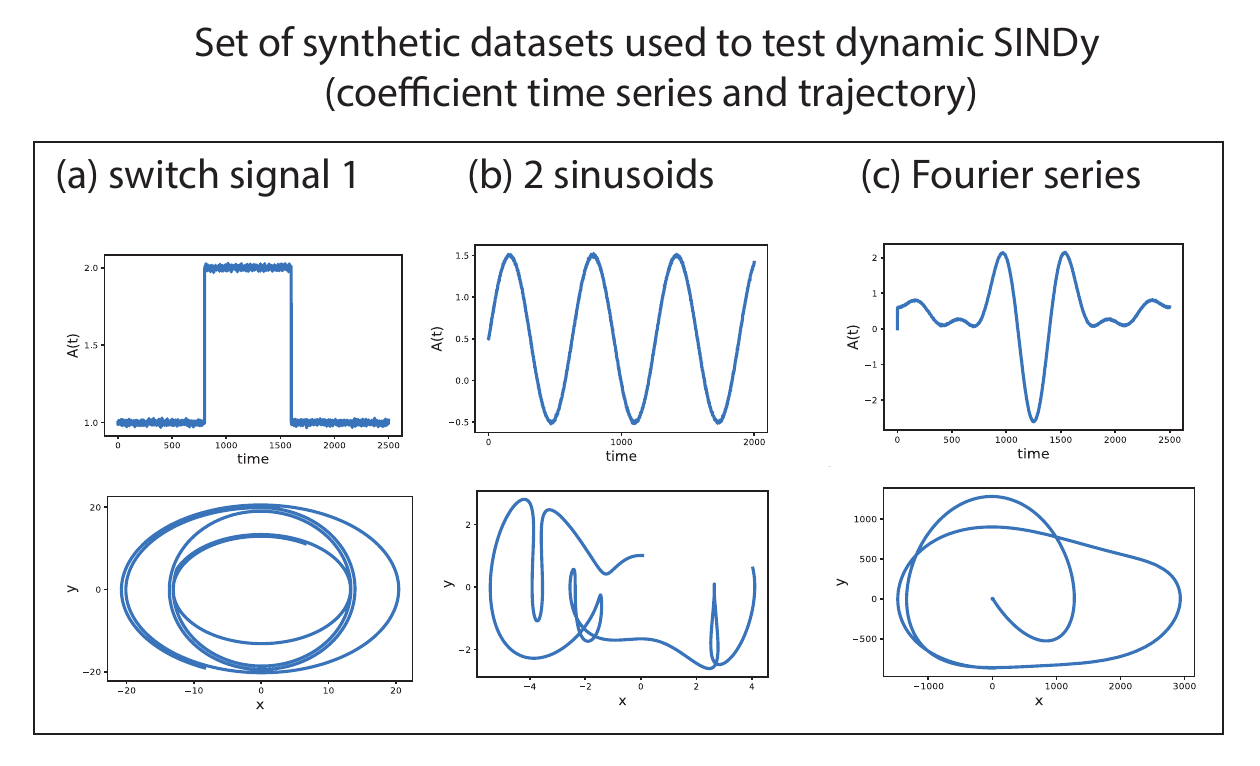}
\caption{Time-varying coefficients (above) and corresponding dynamics (below) for the non-autonomous harmonic oscillator of Eq. (3) (main text). Complementary to Figure 1A (main text)}
\label{suppl_fig2}
\end{figure}

\begin{figure}[h!]
\centering
\includegraphics[width=\textwidth]{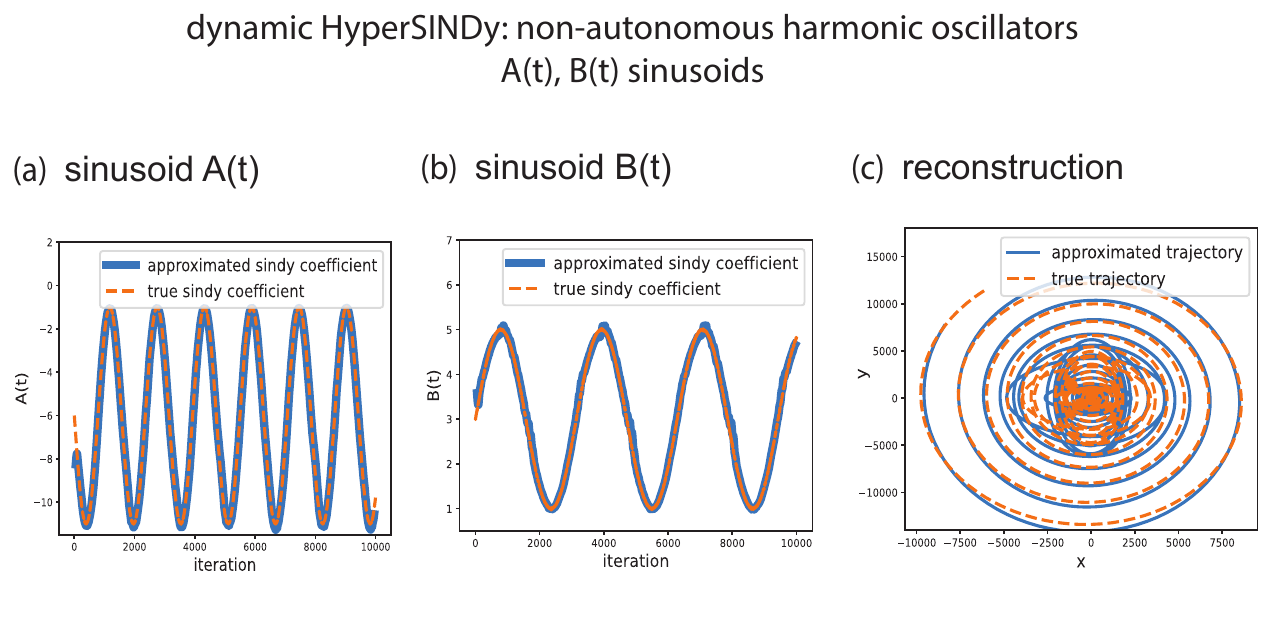}
\caption{Data-driven discovery of sinusoid SINDy coefficients (a and b)and trajectory reconstruction (c) of a non-autonomous harmonic oscillator with dynamic HyperSINDy.}
\label{suppl_fig3}
\end{figure}

\begin{figure}[h!]
\centering
\includegraphics[width=\textwidth]{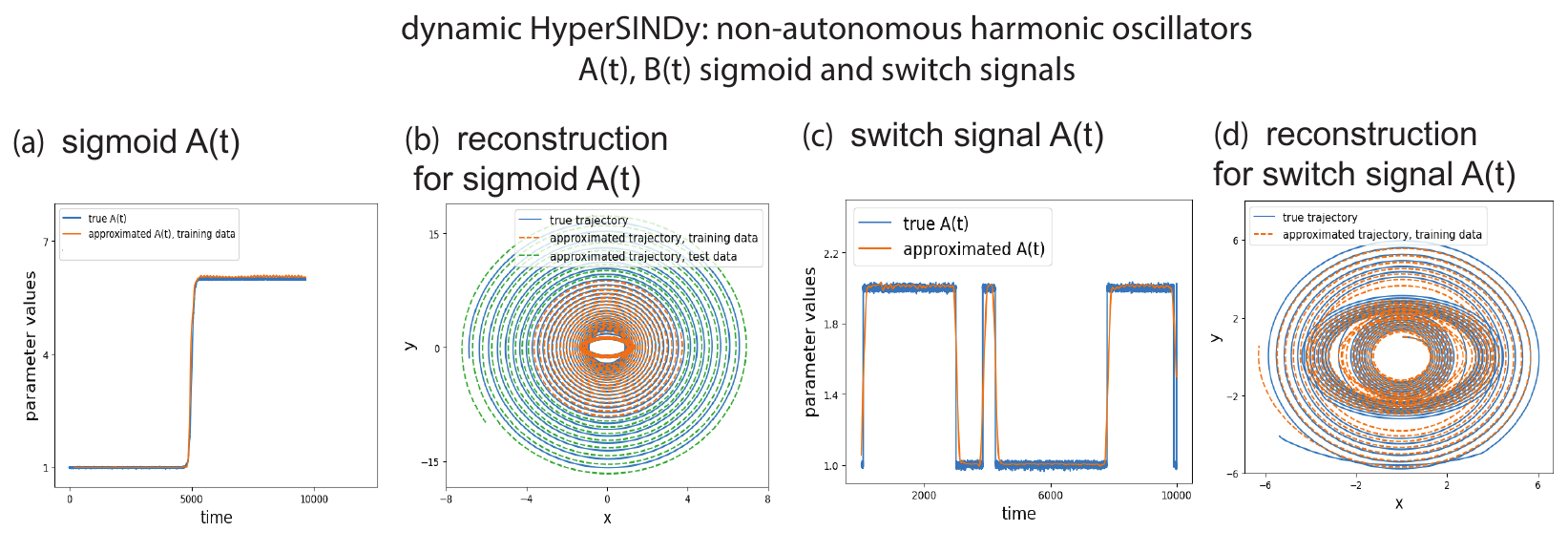}
\caption{Data-driven discovery of sigmoid and switch signal coefficients (a and c) and the corresponding trajectory reconstruction (b and d) of a non-autonomous harmonic oscillator with dynamic HyperSINDy.}
\label{suppl_fig4}
\end{figure}

\begin{figure}[h!]
\centering
\includegraphics[width=\textwidth]{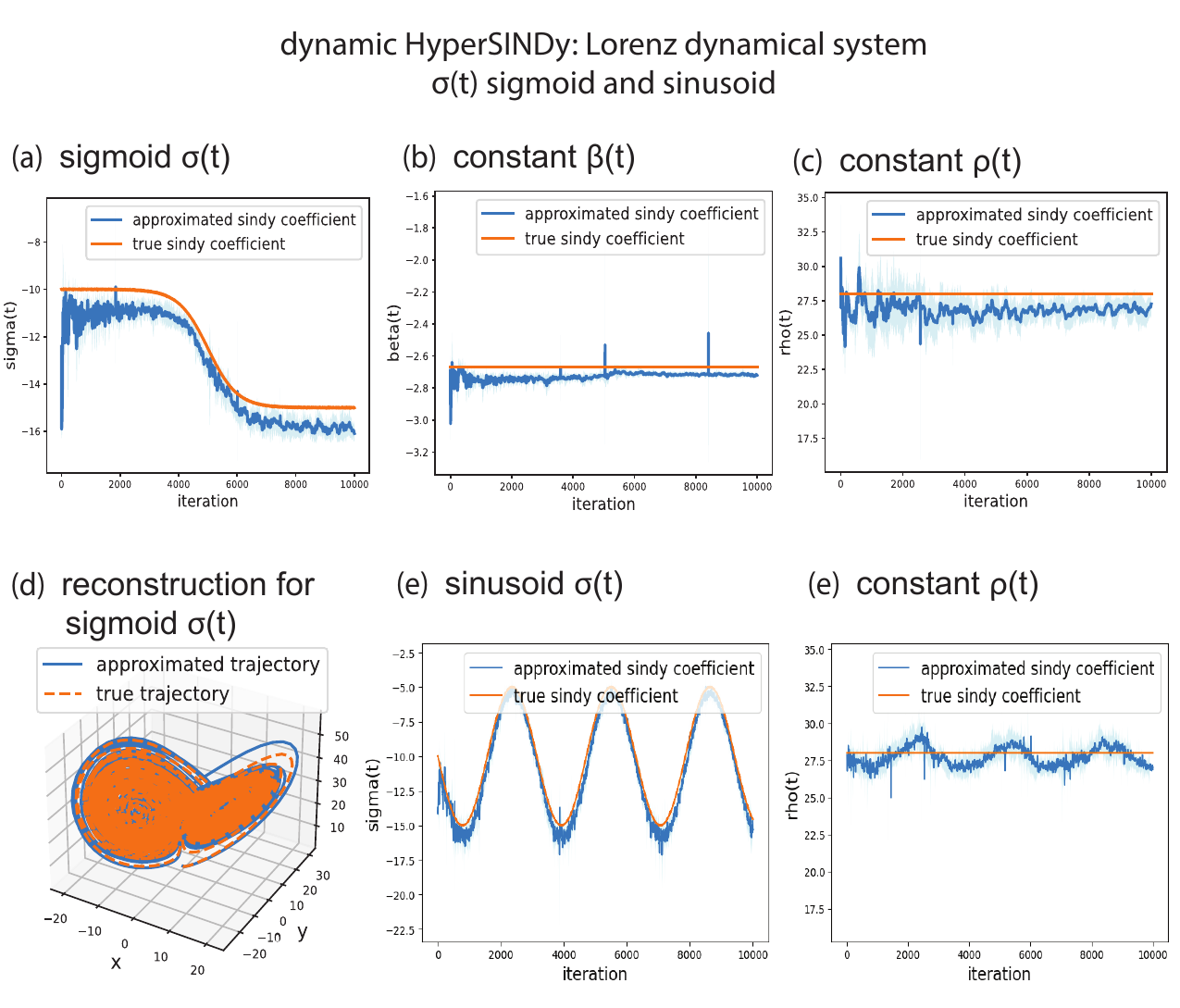}
\caption{Data-driven discovery of sigmoid (a), constant (b, c and e) and sinusoid (e) SINDy coefficients and trajectory reconstruction (d) of the Lorenz dynamics with dynamic HyperSINDy. Time series of coefficients that correspond to constants in the real dynamics sometimes inherit frequency content from the dynamics, as in (e).}
\label{suppl_fig5}
\end{figure}

\begin{figure}[h!]
\centering
\includegraphics[width=\textwidth]{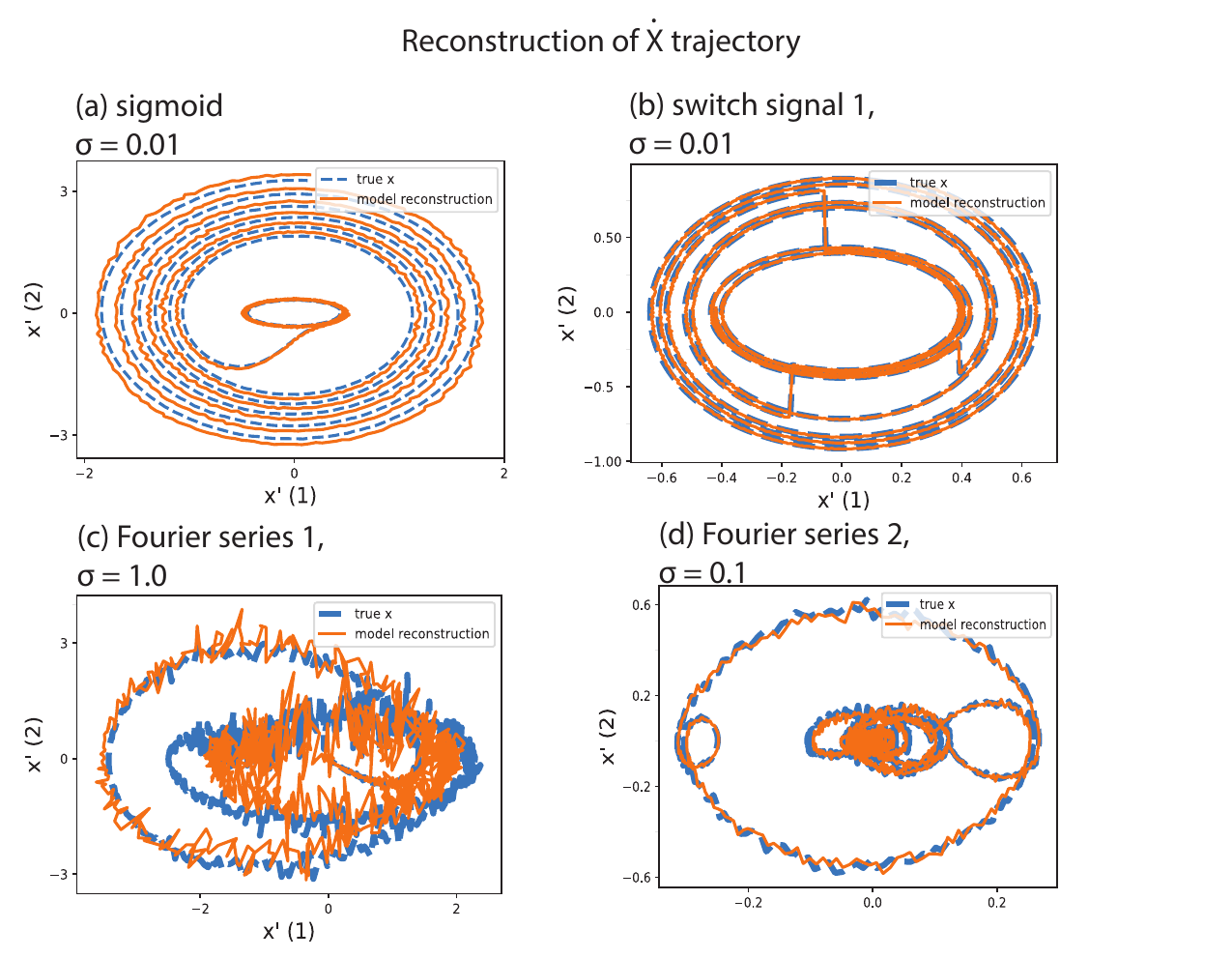}
\caption{Trajectory reconstructions with timeVAE for a non-autonomous harmonic oscillator with different coefficients (sigmoid, switch signal, finite Fourier series) and different levels of noise in the coefficients ($0.01$, $0.1$, $1$).}
\label{suppl_fig6}
\end{figure}